# Every Local Minimum Value Is the Global Minimum Value of Induced Model in Nonconvex Machine Learning

**Kenji Kawaguchi**
*kawaguch@mit.edu*
*MIT, Cambridge, MA 02139, U.S.A.*

**Jiaoyang Huang**
*jiaoyang@math.harvard.edu*
*Harvard University, Cambridge, MA 02138, U.S.A.*

**Leslie Pack Kaelbling**
*lpk@csail.mit.edu*
*MIT, Cambridge, MA 02139, U.S.A.*

**For nonconvex optimization in machine learning, this article proves that every local minimum achieves the globally optimal value of the perturbable gradient basis model at any differentiable point. As a result, nonconvex machine learning is theoretically as supported as convex machine learning with a handcrafted basis in terms of the loss at differentiable local minima, except in the case when a preference is given to the handcrafted basis over the perturbable gradient basis. The proofs of these results are derived under mild assumptions. Accordingly, the proven results are directly applicable to many machine learning models, including practical deep neural networks, without any modification of practical methods. Furthermore, as special cases of our general results, this article improves or complements several state-of-the-art theoretical results on deep neural networks, deep residual networks, and overparameterized deep neural networks with a unified proof technique and novel geometric insights. A special case of our results also contributes to the theoretical foundation of representation learning.**

## 1 Introduction

Deep learning has achieved considerable empirical success in machine learning applications. However, insufficient work has been done on theoretically understanding deep learning, partly because of the nonconvexity and high-dimensionality of the objective functions used to train deep models. In general, theoretical understanding of nonconvex, high-dimensional optimization is challenging. Indeed, finding a global minimum of a general nonconvex function (Murty & Kabadi, 1987) and training certain types of neural networks (Blum & Rivest, 1992) are both NP-hard. Considering the NP-hardness for a general set of relevant problems, it is necessary to use additional assumptions to guarantee efficient global optimality in deep learning. Accordingly, recent theoretical studies have proven global optimality in deep learning by using additional strong assumptions such as linear activation, random activation, semirandom activation, gaussian inputs, single hidden-layer network, and significant overparameterization (Choromanska, Henaff, Mathieu, Ben Arous, & LeCun, 2015; Kawaguchi, 2016; Hardt & Ma, 2017; Nguyen & Hein, 2017, 2018; Brutzkus & Globerson, 2017; Soltanolkotabi, 2017; Ge, Lee, & Ma, 2017; Goel & Klivans, 2017; Zhong, Song, Jain, Bartlett, & Dhillon, 2017; Li & Yuan, 2017; Kawaguchi, Xie, & Song, 2018; Du & Lee, 2018).

A study proving efficient global optimality in deep learning is thus closely related to the search for additional assumptions that might not hold in many practical applications. Toward widely applicable practical theory, we can also ask a different type of question: If standard global optimality requires additional assumptions, then what type of global optimality does not? In other words, instead of searching for additional assumptions to guarantee standard global optimality, we can also search for another type of global optimality under mild assumptions. Furthermore, instead of an arbitrary type of global optimality, it is preferable to develop a general theory of global optimality that not only works under mild assumptions but also produces the previous results with the previous additional assumptions, while predicting new results with future additional assumptions. This type of general theory may help not only to explain when and why an existing machine learning method works but also to predict the types of future methods that will or will not work.

As a step toward this goal, this article proves a series of theoretical results. The major contributions are summarized as follows:

- For nonconvex optimization in machine learning with mild assumptions, we prove that every differentiable local minimum achieves global optimality of the perturbable gradient basis model class. This result is directly applicable to many existing machine learning models, including practical deep learning models, and to new models to be proposed in the future, nonconvex and convex.
- The proposed general theory with a simple and unified proof technique is shown to be able to prove several concrete guarantees that improve or complement several state-of-the-art results.
- In general, the proposed theory allows us to see the effects of the design of models, methods, and assumptions on the optimization landscape through the lens of the global optima of the perturbable gradient basis model class.

Because a local minimum $\theta$ in $\mathbb{R}^{d_\theta}$ only requires the $\theta$ to be locally optimal in $\mathbb{R}^{d_\theta}$, it is nontrivial that the local minimum is guaranteed to achieve the globally optimality in $\mathbb{R}^{d_\theta}$ of the induced perturbable gradient basis model class. The reason we can possibly prove something more than many worst-case results in general nonconvex optimization is that we explicitly take advantage of mild assumptions that commonly hold in machine learning and deep learning. In particular, we assume that an objective function to be optimized is structured with a sum of weighted errors, where each error is an output of composition of a loss function and a function of a hypothesis class. Moreover, we make mild assumptions on the loss function and a hypothesis class, all of which typically hold in practice.

## 2 Preliminaries

This section defines the problem setting and common notation.

**2.1 Problem Description.** Let $x \in \mathcal{X}$ and $y \in \mathcal{Y}$ be an input vector and a target vector, respectively. Define $((x_i, y_i))_{i=1}^m$ as a training data set of size $m$. Let $\theta \in \mathbb{R}^{d_\theta}$ be a parameter vector to be optimized. Let $f(x; \theta) \in \mathbb{R}^{d_y}$ be the output of a model or a hypothesis, and let $\ell : \mathbb{R}^{d_y} \times \mathcal{Y} \to \mathbb{R}_{\geq 0}$ be a loss function. Here, $d_\theta, d_y \in \mathbb{N}_{>0}$. We consider the following standard objective function $L$ to train a model $f(x; \theta)$:

$$L(\theta) = \sum_{i=1}^m \lambda_i \ell(f(x_i; \theta), y_i).$$

This article allows the weights $\lambda_1, \ldots, \lambda_m > 0$ to be arbitrarily fixed. With $\lambda_1 = \cdots = \lambda_m = \frac{1}{m}$, all of our results hold true for the standard average loss $L$ as a special case.

**2.2 Notation.** Because the focus of this article is the optimization of the vector $\theta$, the following notation is convenient: $\ell_y(q) = \ell(q, y)$ and $f_x(q) = f(x; q)$. Then we can write

$$L(\theta) = \sum_{i=1}^m \lambda_i \ell_{y_i}(f_{x_i}(\theta)) = \sum_{i=1}^m \lambda_i (\ell_{y_i} \circ f_{x_i})(\theta).$$

We use the following standard notation for differentiation. Given a scalar-valued or vector-valued function $\varphi : \mathbb{R}^d \to \mathbb{R}^{d'}$ with components $\varphi = (\varphi_1, \ldots, \varphi_{d'})$ and variables $(v_1, \ldots, v_d)$, let $\partial_v \varphi : \mathbb{R}^d \to \mathbb{R}^{d' \times d}$ be the matrix-valued function with each entry $(\partial_v \varphi)_{i,j} = \frac{\partial \varphi_i}{\partial v_j}$. Note that if $\varphi$ is a scalar-valued function, $\partial_v \varphi$ outputs a row vector. In addition, $\partial \varphi = \partial_v \varphi$ if $(v_1, \ldots, v_d)$ are the input variables of $\varphi$. Given a function $\varphi : \mathbb{R}^d \to \mathbb{R}^{d'}$, let $\partial_k \varphi : \mathbb{R}^d \to \mathbb{R}$ be the partial derivative $\partial_k \varphi$ with respect to the $k$th variable of $\varphi$. For the syntax of any differentiation map $\partial$, given functions $\varphi$ and $\zeta$, let $\partial\varphi(\zeta(q)) = (\partial\varphi)(\zeta(q))$ be the (partial) derivative $\partial\varphi$ evaluated at an output $\zeta(q)$ of a function $\zeta$.

Given a matrix $M \in \mathbb{R}^{d \times d'}$, $\text{vec}(M) = [M_{1,1}, \ldots, M_{d,1}, M_{1,2}, \ldots, M_{d,2}, \ldots, M_{1,d'}, \ldots, M_{d,d'}]^T$ represents the standard vectorization of the matrix $M$. Given a set of $n$ matrices or vectors $\{M^{(j)}\}_{j=1}^n$, define $[M^{(j)}]_{j=1}^n = [M^{(1)}, M^{(2)}, \ldots, M^{(n)}]$ to be a block matrix of each column block being $M^{(1)}, M^{(2)}, \ldots, M^{(n)}$. Similarly, given a set $\mathcal{I} = \{i_1, \ldots, i_n\}$ with $(i_1, \ldots, i_n)$ increasing, define $[M^{(j)}]_{j \in \mathcal{I}} = [M^{(i_1)} \cdots M^{(i_n)}]$.

## 3 Nonconvex Optimization Landscapes for Machine Learning

This section shows our first main result that under mild assumptions, every differentiable local minimum achieves the global optimality of the perturbable gradient basis model class.

**3.1 Assumptions.** Given a hypothesis class $f$ and data set, let $\Omega$ be a set of nondifferentiable points $\theta$ as $\Omega = \{\theta \in \mathbb{R}^{d_\theta} : (\exists i \in \{1, \ldots, m\})[f_{x_i} \text{ is not differentiable at } \theta]\}$. Similarly, define $\tilde{\Omega} = \{\theta \in \mathbb{R}^{d_\theta} : (\forall \epsilon > 0)(\exists \theta' \in B(\theta, \epsilon))(\exists i \in \{1, \ldots, m\})[f_{x_i} \text{ is not differentiable at } \theta']\}$. Here, $B(\theta, \epsilon)$ is the open ball with the center $\theta$ and the radius $\epsilon$. In common nondifferentiable models $f$ such as neural networks with rectified linear units (ReLUs) and pooling operations, we have that $\Omega = \tilde{\Omega}$, and the Lebesgue measure of $\Omega(=\tilde{\Omega})$ is zero.

This section uses the following mild assumptions.

**Assumption 1** (Use of Common Loss criteria). For all $i \in \{1, \ldots, m\}$, the function $\ell_{y_i} : q \mapsto \ell(q, y_i) \in \mathbb{R}_{\geq 0}$ is differentiable and convex (e.g., the squared loss, cross-entropy loss, or polynomial hinge loss satisfies this assumption).

**Assumption 2** (Use of Common Model Structures). There exists a function $g : \mathbb{R}^{d_\theta} \to \mathbb{R}^{d_\theta}$ such that $f_{x_i}(\theta) = \sum_{k=1}^{d_\theta} g(\theta)_k \partial_k f_{x_i}(\theta)$ for all $i \in \{1, \ldots, m\}$ and all $\theta \in \mathbb{R}^{d_\theta} \setminus \Omega$.

Assumption 1 is satisfied by simply using common loss criteria that include the squared loss $\ell(q, y) = \|q - y\|_2^2$, cross-entropy loss $\ell(q, y) = -\sum_{k=1}^{d_y} y_k \log \frac{\exp(q_k)}{\sum_{k'} \exp(q_{k'})}$, and smoothed hinge loss $\ell(q, y) = (\max\{0, 1 - yq\})^p$ with $p \geq 2$ (the hinge loss with $d_y = 1$). Although the objective function $L : \theta \mapsto L(\theta)$ used to train a complex machine learning model (e.g., a neural network) is nonconvex in $\theta$, the loss criterion $\ell_{y_i} : q \mapsto \ell(q, y_i)$ is usually convex in $q$. In this article, the cross-entropy loss includes the softmax function, and thus $f_x(\theta)$ is the pre-softmax output of the last layer in related deep learning models.

Assumption 2 is satisfied by simply using a common architecture in deep learning or a classical machine learning model. For example, consider a deep neural network of the form $f_x(\theta) = Wh(x; u) + b$, where $h(x; u)$ is an output of an arbitrary representation at the last hidden layer and $\theta =$ vec($[W, b, u]$). Then assumption 2 holds because $f_{x_i}(\theta) = \sum_{k=1}^{d_\theta} g(\theta)_k \partial_k f_{x_i}(\theta)$, where $g(\theta)_k = \theta_k$ for all $k$ corresponding to the parameters $(W, b)$ in the last layer and $g(\theta)_k = 0$ for all other $k$ corresponding to $u$. In general, because $g$ is a function of $\theta$, assumption 2 is easily satisfiable. Assumption 2 does not require the model $f(x; \theta)$ to be linear in $\theta$ or $x$.

Note that we allow the nondifferentiable points to exist in $L(\theta)$; for example, the use of ReLU is allowed. For a nonconvex and nondifferentiable function, we can still have first-order and second-order necessary conditions of local minima (e.g., Rockafellar & Wets, 2009, theorem 13.24). However, subdifferential calculus of a nonconvex function requires careful treatment at nondifferentiable points (see Rockafellar & Wets, 2009; Kakade & Lee, 2018; Davis, Drusvyatskiy, Kakade, & Lee, 2019), and deriving guarantees at nondifferentiable points is left to a future study.

**3.2 Theory for Critical Points.** Before presenting the first main result, this section provides a simpler result for critical points to illustrate the ideas behind the main result for local minima. We define the (theoretical) objective function $L_\theta$ of the gradient basis model class as

$$L_\theta(\alpha) = \sum_{i=1}^{m} \lambda_i \ell \left( f_\theta(x_i; \alpha), y_i \right),$$

where $\{ f_\theta(x_i; \alpha) = \sum_{k=1}^{d_\theta} \alpha_k \partial_k f_{x_i}(\theta) : \alpha \in \mathbb{R}^{d_\theta} \}$ is the induced gradient basis model class. The following theorem shows that every differentiable critical point of our original objective $L$ (including every differentiable local minimum and saddle point) achieves the global minimum value of $L_\theta$. The complete proofs of all the theoretical results are presented in appendix A.

**Theorem 1.** *Let assumptions 1 and 2 hold. Then for any critical point $\theta \in (\mathbb{R}^{d_\theta} \setminus \Omega)$ of $L$, the following holds:*

$$L(\theta) = \inf_{\alpha \in \mathbb{R}^{d_\theta}} L_\theta(\alpha).$$

An important aspect in theorem 1 is that $L_\theta$ on the right-hand side is convex, while $L$ on the left-hand side can be nonconvex or convex. Here, following convention, $\inf S$ is defined to be the infimum of a subset $S$ of $\overline{\mathbb{R}}$ (the set of affinely extended real numbers); that is, if $S$ has no lower bound, $\inf S = -\infty$ and $\inf \emptyset = \infty$. Note that theorem 1 vacuously holds true if there is no critical point for $L$. To guarantee the existence of a minimizer in a (nonempty) subspace $S \subseteq \mathbb{R}^{d_\theta}$ for $L$ (or $L_\theta$), a classical proof requires two conditions: a lower semicontinuity of $L$ (or $L_\theta$) and the existence of a $q \in S$ for which the set $\{q' \in S : L(q') \leq L(q)\}$ (or $\{q' \in S : L_\theta(q') \leq L_\theta(q)\}$) is compact (see Bertsekas, 1999, for different conditions).

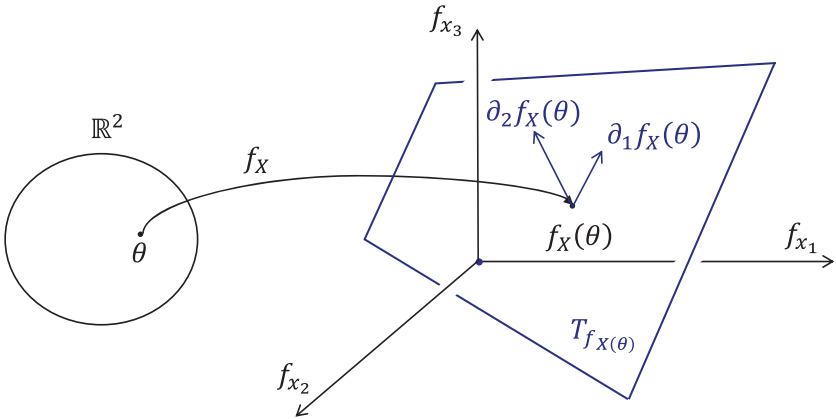

Figure 1: Illustration of gradient basis model class and theorem 1 with $\theta \in \mathbb{R}^2$ and $f_X(\theta) \in \mathbb{R}^3$ ($d_y = 1$). Theorem 1 translates the local condition of $\theta$ in the parameter space $\mathbb{R}^2$ (on the left) to the global optimality in the output space $\mathbb{R}^3$ (on the right). The subspace $T_{f_X(\theta)}$ is the space of the outputs of the gradient basis model class. Theorem 1 states that $f_X(\theta)$ is globally optimal in the subspace as $f_X(\theta) \in \mathrm{argmin}_{\mathbf{f} \in T_{f_X(\theta)}} \mathrm{dist}(\mathbf{f}, \mathbf{y})$ for any differentiable critical point $\theta$ of $L$.

*3.2.1 Geometric View.* This section presents the geometric interpretation of theorem 1 that provides an intuitive yet formal description of gradient basis model class. Figure 1 illustrates the gradient basis model class and theorem 1 with $\theta \in \mathbb{R}^2$ and $f_X(\theta) \in \mathbb{R}^3$. Here, we consider the following map from the parameter space to the concatenation of the output of the model at $x_1, x_2, \ldots, x_m$:

$$f_X : \theta \in \mathbb{R}^{d_\theta} \mapsto (f_{x_1}(\theta)^\top, f_{x_2}(\theta)^\top, \ldots, f_{x_m}(\theta)^\top)^\top \in \mathbb{R}^{md_y}.$$

In the output space $\mathbb{R}^{md_y}$ of $f_X$, the objective function $L$ induces the notion of distance from the target vector $\mathbf{y} = (y_1^\top, \ldots, y_m^\top)^\top \in \mathbb{R}^{md_y}$ to a vector $\mathbf{f} = (\mathbf{f}_1^\top, \ldots, \mathbf{f}_m^\top)^\top \in \mathbb{R}^{md_y}$ as

$$\mathrm{dist}(\mathbf{f}, \mathbf{y}) = \sum_{i=1}^{m} \lambda_i \ell(\mathbf{f}_i, y_i).$$

We consider the affine subspace $T_{f_X(\theta)}$ of $\mathbb{R}^{md_y}$ that passes through the point $f_X(\theta)$ and is spanned by the set of vectors $\{\partial_1 f_X(\theta), \ldots, \partial_{d_\theta} f_X(\theta)\}$,

$$T_{f_X(\theta)} = \mathrm{span}(\{\partial_1 f_X(\theta), \ldots, \partial_{d_\theta} f_X(\theta)\}) + \{f_X(\theta)\},$$

where the sum of the two sets represents the Minkowski sum of the sets.

Then the subspace $T_{f_X(\theta)}$ is the space of the outputs of the gradient basis model class in general beyond the low-dimensional illustration. This is because by assumption 2, for any given $\theta$,

$$\begin{aligned}
T_{f_X(\theta)} &= \left\{ \sum_{k=1}^{d_\theta} (g(\theta)_k + \alpha_k) \partial_k f_X(\theta) : \alpha \in \mathbb{R}^{d_\theta} \right\} \\
&= \left\{ \sum_{k=1}^{d_\theta} \alpha_k \partial_k f_X(\theta) : \alpha \in \mathbb{R}^{d_\theta} \right\},
\end{aligned} \qquad (3.1)$$

and $\sum_{k=1}^{d_\theta} \alpha_k \partial_k f_X(\theta) = (f_\theta(x_1;\alpha)^\top, \ldots, f_\theta(x_m;\alpha)^\top)^\top$. In other words, $T_{f_X(\theta)}$ = $\mathrm{span}(\{\partial_1 f_X(\theta), \ldots, \partial_{d_\theta} f_X(\theta)\}) \ni (f_\theta(x_1;\alpha)^\top, \ldots, f_\theta(x_m;\alpha)^\top)^\top$.

Therefore, in general, theorem 1 states that under assumptions 1 and 2, $f_X(\theta)$ is globally optimal in the subspace $T_{f_X(\theta)}$ as

$$f_X(\theta) \in \underset{\mathbf{f} \in T_{f_X(\theta)}}{\mathrm{argmin}}\ \mathrm{dist}(\mathbf{f}, \mathbf{y}),$$

for any differentiable critical point $\theta$ of $L$. Theorem 1 concludes this global optimality in the affine subspace of the output space based on the local condition in the parameter space (i.e., differentiable critical point). A key idea behind theorem 1 is to consider the map between the parameter space and the output space, which enables us to take advantage of assumptions 1 and 2.

Figure 2 illustrates the gradient basis model class and theorem 1 with a union of manifolds and a tangent space. Under the constant rank condition, the image of the map $f_X$ locally forms a single manifold. More precisely, if there exists a small neighborhood $U(\theta)$ of $\theta$ such that $f_X$ is differentiable in $U(\theta)$ and $\mathrm{rank}(\partial f_X(\theta')) = r$ is constant with some $r$ for all $\theta' \in U(\theta)$ (the constant rank condition), then the rank theorem states that the image $f_X(U(\theta))$ is a manifold of dimension $r$ (Lee, 2013, theorem 4.12). We note that the rank map $\theta \mapsto \mathrm{rank}(\partial f_X(\theta))$ is lower semicontinuous (i.e., if $\mathrm{rank}(\partial f_X(\theta)) = r$, then there exists a neighborhood $U(\theta)$ of $\theta$ such that $\mathrm{rank}(\partial f_X(\theta')) \geq r$ for any $\theta' \in U(\theta)$). Therefore, if $\partial f_X(\theta)$ at $\theta$ has the maximum rank in a small neighborhood of $\theta$, then the constant rank condition is satisfied.

For points $\theta$ where the constant rank condition is violated, the image of the map $f_X$ is no longer a single manifold. However, locally it decomposes as a union of finitely many manifolds. More precisely, if there exists a small neighborhood $U(\theta)$ of $\theta$ such that $f_X$ is analytic over $U(\theta)$ (this condition is satisfied for commonly used activation functions such as ReLU, sigmoid, and hyperbolic tangent at any differentiable point), then the image $f_X(U(\theta))$ admits a locally finite partition $\mathcal{M}$ into connected submanifolds such that whenever $M \neq M' \in \mathcal{M}$ with $\bar{M} \cap M' \neq \emptyset$ ($\bar{M}$ is the closure of $M$), we have

$$M' \subset \bar{M}, \quad \dim(M') < \dim(M).$$

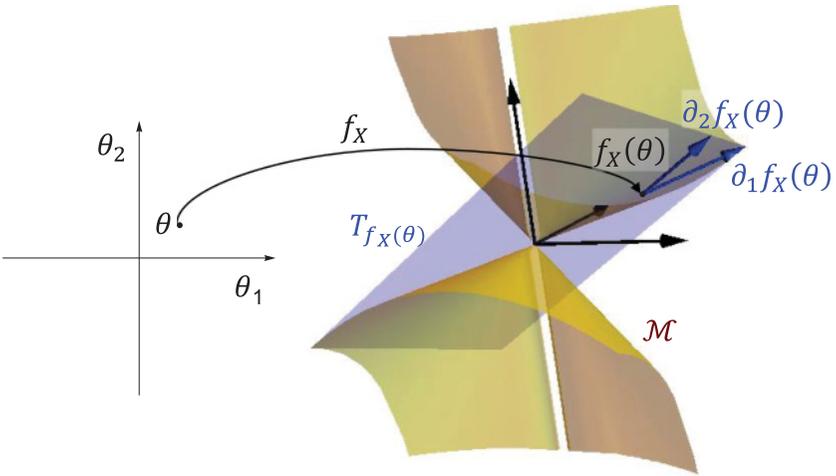

Figure 2: Illustration of gradient basis model class and theorem 1 with manifold and tangent space. The space $\mathbb{R}^2 \ni \theta$ on the left is the parameter space, and the space $\mathbb{R}^3 \ni f_X(\theta)$ on the right is the output space. The surface $\mathcal{M} \subset \mathbb{R}^3$ on the right is the image of $f_X$, which is a union of finitely many manifolds. The tangent space $T_{f_X(\theta)}$ is the space of the outputs of the gradient basis model class. Theorem 1 states that if $\theta$ is a differentiable critical point of $L$, then $f_X(\theta)$ is globally optimal in the tangent space $T_{f_X(\theta)}$.

See Hardt (1975) for the proof.

If the point $\theta$ satisfies the constant rank condition, then $T_{f_X(\theta)}$ is exactly the tangent space of the manifold formed by the image $f_X(U(\theta))$. Otherwise, locally the image decomposes into a finite union $\mathcal{M}$ of submanifolds. In this case, $T_{f_X(\theta)}$ belongs to the span of the tangent space of those manifolds in $\mathcal{M}$ as

$$T_{f_X(\theta)} \subset \{T_p M : p = f_X(\theta), M \in \mathcal{M}\},$$

where $T_p M$ is the tangent space of the manifold $M$ at the point $p$.

*3.2.2 Examples.* In this section, we show through examples that theorem 1 generalizes the previous results in special cases while providing new theoretical insights based on the gradient basis model class and its geometric view. In the following, whenever the form of $f$ is specified, we require only assumption 1 because assumption 2 is automatically satisfied by a given $f$.

For classical machine learning models, example 1 shows that the gradient basis model class is indeed equivalent to a given model class. From the geometric view, this means that for any $\theta$, the tangent space $T_{f_X(\theta)}$ is equal to the whole image $\mathcal{M}$ of $f_X$ (i.e., $T_{f_X(\theta)}$ does not depend on $\theta$). This reduces theorem 1 to the statement that every critical point of $L$ is a global minimum of $L$.

**Example 1: Classical Machine Learning Models.** For any basis function model $f(x; \theta) = \sum_{k=1}^{d_\theta} \theta_k \phi(x)_k$ in classical machine learning with any fixed feature map $\phi : \mathcal{X} \to \mathbb{R}^{d_\theta}$, we have that $f_\theta(x; \alpha) = f(x; \alpha)$, and hence $\inf_{\theta \in \mathbb{R}^{d_\theta}} L(\theta) = \inf_{\alpha \in \mathbb{R}^{d_\theta}} L_\theta(\alpha)$, as well as $\Omega = \emptyset$. In other words, in this special case, theorem 1 states that every critical point of $L$ is a global minimum of $L$. Here, we do not assume that a critical point or a global minimum exists or can be attainable. Instead, the statement logically means that if a point is a critical point, then the point is a global minimum. This type of statement vacuously holds true if there is no critical point.

For overparameterized deep neural networks, example 2 shows that the induced gradient basis model class is highly expressive such that it must contain the globally optimal model of a given model class of deep neural networks. In this example, the tangent space $T_{f_X(\theta)}$ is equal to the whole output space $\mathbb{R}^{md_y}$. This reduces theorem 1 to the statement that every critical point of $L$ is a global minimum of $L$ for overparameterized deep neural networks.

Intuitively, in Figure 1 or 2, we can increase the number of parameters and raise the number of partial derivatives $\partial_k f_X(\theta)$ in order to increase the dimensionality of the tangent space $T_{f_X(\theta)}$ so that $T_{f_X(\theta)} = \mathbb{R}^{md_y}$. This is indeed what happens in example 2, as well as in the previous studies of significantly overparameterized deep neural networks (Allen-Zhu, Li, & Song, 2018; Du, Lee, Li, Wang, & Zhai, 2018; Zou et al., 2018). In the previous studies, the significant overparameterization is required so that the tangent space $T_{f_X(\theta)}$ does not change from the initial tangent space $T_{f_X(\theta^{(0)})} = \mathbb{R}^{md_y}$ during training. Thus, theorem 1, with its geometric view, provides the novel algebraic and geometric insights into the results of the previous studies and the reason why overparameterized deep neural networks are easy to be optimized despite nonconvexity.

**Example 2: Overparameterized Deep Neural Networks.** Theorem 1 implies that every critical point (and every local minimum) is a global minimum for sufficiently overparameterized deep neural networks. Let $n$ be the number of units in each layer of a fully connected feedforward deep neural network. Let us consider a significant overparameterization such that $n \geq m$. Let us write a fully connected feedforward deep neural network with the trainable parameters $(\theta, u)$ by $f(x; \theta) = W\phi(x; u)$, where $W \in \mathbb{R}^{d_y \times n}$ is the weight matrix in the last layer, $\theta = \text{vec}(W)$, $u$ contains the rest of the parameters, and $\phi(x; u)$ is the output of the last hidden layer. Denote $x_i = [(x_i^{(\text{raw})})^\top, 1]^\top$ to contain the constant term to account for the bias term in the first layer. Assume that the input samples are normalized as $\|x_i^{(\text{raw})}\|_2 = 1$ for all $i \in \{1, \ldots, m\}$ and distinct as $(x_i^{(\text{raw})})^\top x_{i'}^{(\text{raw})} < 1 - \delta$ with some $\delta > 0$ for all $i' \neq i$. Assume that the activation functions are ReLU activation functions. Then we can efficiently set $u$ to guarantee $\text{rank}([\phi(x_i; u)]_{i=1}^m) \geq m$ (e.g., by choosing $u$ to make each unit of the last layer to be active only for each sample $x_i$).[1] Theorem 1 implies that every critical point $\theta$ with this $u$ is a global minimum of the whole set of trainable parameters $(\theta, u)$ because $\inf_\alpha L_\theta(\alpha) = \inf_{f_1, \ldots, f_m} \sum_{i=1}^m \lambda_i \ell(f_i, y_i)$ (with assumption 1).

For deep neural networks, example 3 shows that standard networks have the global optimality guarantee with respect to the representation learned at the last layer, and skip connections further ensure the global optimality with respect to the representation learned at each hidden layer. This is because adding the skip connections incurs new partial derivatives $\{\partial_k f_X(\theta)\}_k$ that span the tangent space containing the output of the best model with the corresponding learned representation.

**Example 3: Deep Neural Networks and Learned Representations.** Consider a feedforward deep neural network, and let $\mathcal{I}^{(\text{skip})} \subseteq \{1, \ldots, H\}$ be the set of indices such that there exists a skip connection from the $(l-1)$th layer to the last layer for all $l \in \mathcal{I}^{(\text{skip})}$; that is, in this example,

$$f(x; \theta) = \sum_{l \in \mathcal{I}^{(\text{skip})}} W^{(l+1)} h^{(l)}(x; u),$$

where $\theta = \text{vec}([[W^{(l+1)}]_{l \in \mathcal{I}^{(\text{skip})}}, u]) \in \mathbb{R}^{d_\theta}$ with $W^{(l+1)} \in \mathbb{R}^{d_y \times d_l}$ and $u \in \mathbb{R}^{d_u}$.

The conclusion in this example holds for standard deep neural networks without skip connections too, since we always have $H \in \mathcal{I}^{(\text{skip})}$ for standard deep neural networks. Let assumption 1 hold. Then theorem 1 implies that for any critical point $\theta \in (\mathbb{R}^{d_\theta} \setminus \Omega)$ of $L$, the following holds:

$$L(\theta) = \inf_{\alpha \in \mathbb{R}^{d_\theta}} L_\theta^{(\text{skip})}(\alpha),$$

---

[1] For example, choose the first layer's weight matrix $W^{(1)}$ such that for all $i \in \{1, \ldots, m\}$, $(W^{(1)} x_i)_i > 0$ and $(W^{(1)} x_i)_{i'} \leq 0$ for all $i' \neq i$. This can be achieved by choosing the $i$th row of $W^{(1)}$ to be $[(x_i^{(\text{raw})})^\top, \epsilon - 1]$ with $0 < \epsilon \leq \delta$ for $i \leq m$. Then choose the weight matrices for the $l$th layer for all $l \geq 2$ such that for all $j$, $W_{j,j}^{(l)} \neq 0$ and $W_{j',j}^{(l)} = 0$ for all $j' \neq j$. This guarantees $\text{rank}([\phi(x_i; u)]_{i=1}^m) \geq m$.

where

$$L_\theta^{(\text{skip})}(\alpha) = \sum_{i=1}^m \lambda_i \ell_{y_i}\left(\sum_{l \in \mathcal{I}^{(\text{skip})}} \alpha_w^{(l+1)} h^{(l)}(x_i; u) + \sum_{k=1}^{d_u} (\alpha_u)_k \partial_{u_k} f_{x_i}(\theta)\right),$$

with $\alpha = \text{vec}([[\alpha^{(l+1)}]_{l \in \mathcal{I}^{(\text{skip})}}, \alpha_u]) \in \mathbb{R}^{d_\theta}$ with $\alpha^{(l+1)} \in \mathbb{R}^{d_y \times d_l}$ and $\alpha_u \in \mathbb{R}^{d_u}$. This is because $f(x; \theta) = (\partial_{\text{vec}(W^{(H+1)})} f(x; \theta)) \text{vec}(W^{(H+1)})$, and thus assumption 2 is automatically satisfied. Here, $h^{(l)}(x_i; u)$ is the representation learned at the $l$-layer. Therefore, $\inf_{\alpha \in \mathbb{R}^{d_\theta}} L_\theta^{(\text{skip})}(\alpha)$ is at most the global minimum value of the basis models with the learned representations of the last layer and all hidden layers with the skip connections.

**3.3 Theory for Local Minima.** We are now ready to present our first main result. We define the (theoretical) objective function $\tilde{L}_\theta$ of the perturbable gradient basis model class as

$$\tilde{L}_\theta(\alpha, \epsilon, S) = \sum_{i=1}^m \lambda_i \ell(\tilde{f}_\theta(x_i; \alpha, \epsilon, S), y_i),$$

where $\tilde{f}_\theta(x_i; \alpha, \epsilon, S)$ is a perturbed gradient basis model defined as

$$\tilde{f}_\theta(x_i; \alpha, \epsilon, S) = \sum_{k=1}^{d_\theta} \sum_{j=1}^{|S|} \alpha_{k,j} \partial_k f_{x_i}(\theta + \epsilon S_j).$$

Here, $S$ is a finite set of vectors $S_1, \ldots, S_{|S|} \in \mathbb{R}^{d_\theta}$ and $\alpha \in \mathbb{R}^{d_\theta \times |S|}$. Let $\mathcal{V}[\theta, \epsilon]$ be the set of all vectors $v \in \mathbb{R}^{d_\theta}$ such that $\|v\|_2 \leq 1$ and $f_{x_i}(\theta + \epsilon v) = f_{x_i}(\theta)$ for any $i \in \{1, \ldots, m\}$. Let $S \subseteq_{\text{fin}} S'$ denote a finite subset $S$ of a set $S'$. For an $S_j \in \mathcal{V}[\theta, \epsilon]$, we have $f_{x_i}(\theta + \epsilon S_j) = f_{x_i}(\theta)$, but it is possible to have $\partial_k f_{x_i}(\theta + \epsilon S_j) \neq \partial_k f_{x_i}(\theta)$. This enables the greater expressivity of $\tilde{f}_\theta(x_i; \alpha, \epsilon, S)$ with a $S \subseteq_{\text{fin}} \mathcal{V}[\theta, \epsilon]$ when compared with $f_\theta(x_i; \alpha)$.

The following theorem shows that every differentiable local minimum of $L$ achieves the global minimum value of $\tilde{L}_\theta$:

**Theorem 2.** *Let assumptions 1 and 2 hold. Then, for any local minimum $\theta \in (\mathbb{R}^{d_\theta} \setminus \tilde{\Omega})$ of $L$, the following holds: there exists $\epsilon_0 > 0$ such that for any $\epsilon \in [0, \epsilon_0)$,*

$$L(\theta) = \inf_{\substack{S \subseteq_{\text{fin}} \mathcal{V}[\theta, \epsilon], \\ \alpha \in \mathbb{R}^{d_\theta \times |S|}}} \tilde{L}_\theta(\alpha, \epsilon, S). \tag{3.2}$$

To understand the relationship between theorems 1 and 2, let us consider the following general inequalities: for any $\theta \in (\mathbb{R}^{d_\theta} \setminus \tilde{\Omega})$ with $\epsilon \geq 0$ being sufficiently small,

$$L(\theta) \geq \inf_{\alpha \in \mathbb{R}^{d_\theta}} L_\theta(\alpha) \geq \inf_{\substack{S \subseteq_{fin} \mathcal{V}[\theta,\epsilon], \\ \alpha \in \mathbb{R}^{d_\theta \times |S|}}} \tilde{L}_\theta(\alpha, \epsilon, S).$$

Here, whereas theorem 1 states that the first inequality becomes equality as $L(\theta) = \inf_{\alpha \in \mathbb{R}^{d_\theta}} L_\theta(\alpha)$ at every differentiable critical point, theorem 2 states that both inequalities become equality as

$$L(\theta) = \inf_{\alpha \in \mathbb{R}^{d_\theta}} L_\theta(\alpha) = \inf_{\substack{S \subseteq_{fin} \mathcal{V}[\theta,\epsilon], \\ \alpha \in \mathbb{R}^{d_\theta \times |S|}}} \tilde{L}_\theta(\alpha, \epsilon, S)$$

at every differentiable local minimum.

From theorem 1 to theorem 2, the power of increasing the number of parameters (including overparameterization) is further improved. The right-hand side in equation 3.2 is the global minimum value over the variables $S \subseteq_{fin} \mathcal{V}[\theta,\epsilon]$ and $\alpha \in \mathbb{R}^{d_\theta \times |S|}$. Here, as $d_\theta$ increases, we may obtain the global minimum value of a larger search space $\mathbb{R}^{d_\theta \times |S|}$, which is similar to theorem 1. A concern in theorem 1 is that as $d_\theta$ increases, we may also significantly increase the redundancy among the elements in $\{\partial_k f_{x_i}(\theta)\}_{k=1}^{d_\theta}$. Although this remains a valid concern, theorem 2 allows us to break the redundancy by the globally optimal $S \subseteq_{fin} \mathcal{V}[\theta,\epsilon]$ to some degree.

For example, consider $f(x;\theta) = g(W^{(l)}h^{(l)}(x;u);u)$, which represents a deep neural network, with some $l$th-layer output $h^{(l)}(x;u) \in \mathbb{R}^{d_l}$, a trainable weight matrix $W^{(l)}$, and an arbitrary function $g$ to compute the rest of the forward pass. Here, $\theta = \text{vec}([W^{(l)}, u])$. Let $h^{(l)}(X;u) = [h^{(l)}(x_i;u)]_{i=1}^m \in \mathbb{R}^{d_l \times m}$ and, similarly, $f(X;\theta) = g(W^{(l)}h^{(l)}(X;u);u) \in \mathbb{R}^{d_y \times m}$. Then, all vectors $v$ corresponding to any elements in the left null space of $h^{(l)}(X;u)$ are in $\mathcal{V}[\theta,\epsilon]$ (i.e., $v_k = 0$ for all $k$ corresponding to $u$ and the rest of $v_k$ is set to perturb $W^{(l)}$ by an element in the left null space). Thus, as the redundancy increases such that the dimension of the left null space of $h^{(l)}(X;u)$ increases, we have a larger space of $\mathcal{V}[\theta,\epsilon]$, for which a global minimum value is guaranteed at a local minimum.

3.3.1 Geometric View. This section presents the geometric interpretation of the perturbable gradient basis model class and theorem 2. Figure 3 illustrates the perturbable gradient basis model class and theorem 2 with $\theta \in \mathbb{R}^2$ and $f_X(\theta) \in \mathbb{R}^3$. Figure 4 illustrates them with a union of manifolds and tangent spaces at a singular point. Given a $\epsilon$ ($\leq \epsilon_0$), define the affine subspace $\tilde{T}_{f_X(\theta)}$ of the output space $\mathbb{R}^{md_y}$ by

$$\tilde{T}_{f_X(\theta)} = \text{span}(\{\mathbf{f} \in \mathbb{R}^{md_y} : (\exists v \in \mathcal{V}[\theta,\epsilon])[\mathbf{f} \in T_{f_X(\theta+\epsilon v)}]\}).$$

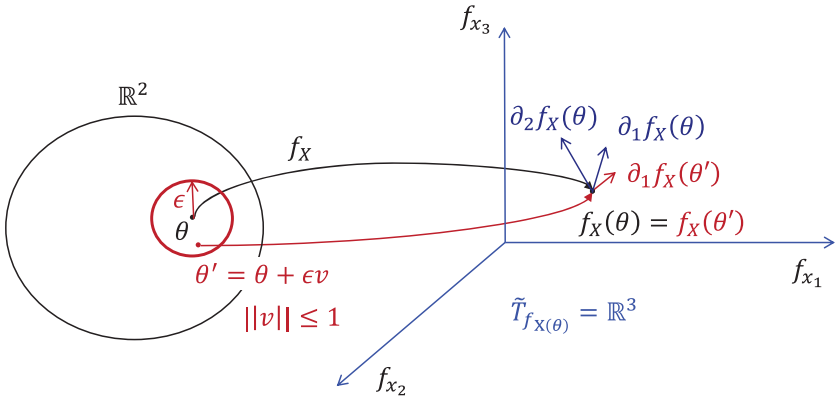

Figure 3: Illustration of perturbable gradient basis model class and theorem 2 with $\theta \in \mathbb{R}^2$ and $f_X(\theta) \in \mathbb{R}^3$ ($d_y = 1$). Theorem 2 translates the local condition of $\theta$ in the parameter space $\mathbb{R}^2$ (on the left) to the global optimality in the output space $\mathbb{R}^3$ (on the right). The subspace $\tilde{T}_{f_X(\theta)}$ is the space of the outputs of the perturbable gradient basis model class. Theorem 2 states that $f_X(\theta)$ is globally optimal in the subspace as $f_X(\theta) \in \text{argmin}_{\mathbf{f} \in \tilde{T}_{f_X(\theta)}} \text{dist}(\mathbf{f}, \mathbf{y})$ for any differentiable local minima $\theta$ of $L$. In this example, $\tilde{T}_{f_X(\theta)}$ is the whole output space $\mathbb{R}^3$, while $T_{f_X(\theta)}$ is not, illustrating the advantage of the perturbable gradient basis over the gradient basis. Since $\tilde{T}_{f_X(\theta)} = \mathbb{R}^3$, $f_X(\theta)$ must be globally optimal in the whole output space $\mathbb{R}^3$.

Then the subspace $\tilde{T}_{f_X(\theta)}$ is the space of the outputs of the perturbable gradient basis model class in general beyond the low-dimensional illustration (this follows equation 3.1 and the definition of the perturbable gradient basis model). Therefore, in general, theorem 2 states that under assumptions 1 and 2, $f_X(\theta)$ is globally optimal in the subspace $\tilde{T}_{f_X(\theta)}$ as

$$f_X(\theta) \in \underset{\mathbf{f} \in \tilde{T}_{f_X(\theta)}}{\text{argmin}} \ \text{dist}(\mathbf{f}, \mathbf{y})$$

for any differentiable local minima $\theta$ of $L$. Theorem 2 concludes the global optimality in the affine subspace of the output space based on the local condition in the parameter space—that is, differentiable local minima. Here, a (differentiable) local minimum $\theta$ is required to be optimal only in an arbitrarily small local neighborhood in the parameter space, and yet $f_X(\theta)$ is guaranteed to be globally optimal in the affine subspace of the output space. This illuminates the fact that nonconvex optimization in machine learning has a particular structure beyond general nonconvex optimization.

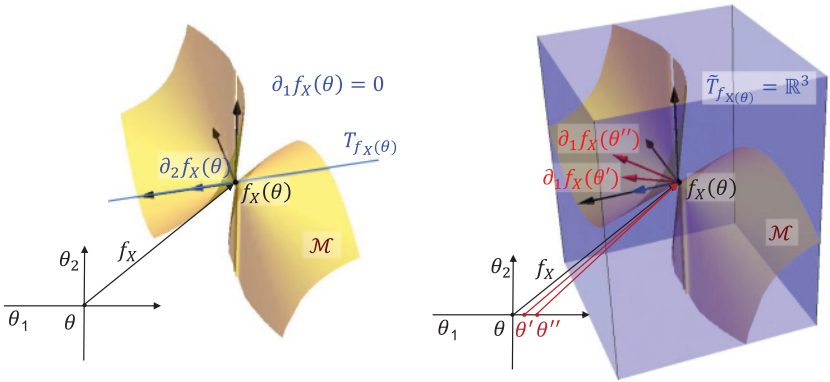

Figure 4: Illustration of perturbable gradient basis model class and theorem 2 with manifold and tangent space at a singular point. The surface $\mathcal{M} \subset \mathbb{R}^3$ is the image of $f_X$, which is a union of finitely many manifolds. The line $T_{f_X(\theta)}$ on the left panel is the space of the outputs of the gradient basis model class. The whole space $\tilde{T}_{f_X(\theta)} = \mathbb{R}^3$ on the right panel is the space of the outputs of the perturbable gradient basis model class. The space $\tilde{T}_{f_X(\theta)}$ is the span of the set of the vectors in the tangent spaces $T_{f_X(\theta)}$, $T_{f_X(\theta')}$, and $T_{f_X(\theta'')}$. Theorem 2 states that if $\theta$ is a differentiable local minimum of $L$, then $f_X(\theta)$ is globally optimal in the space $\tilde{T}_{f_X(\theta)}$.

## 4 Applications to Deep Neural Networks

The previous section showed that all local minima achieve the global optimality of the perturbable gradient basis model class with several direct consequences for special cases. In this section, as consequences of theorem 2, we complement or improve the state-of-the-art results in the literature.

**4.1 Example: ResNets.** As an example of theorem 2, we set $f$ to be the function of a certain type of residual networks (ResNets) that Shamir (2018) studied. That is, both Shamir (2018) and this section set $f$ as

$$f(x; \theta) = W(x + Rz(x; u)), \quad (4.1)$$

where $\theta = \text{vec}([W, R, u]) \in \mathbb{R}^{d_\theta}$ with $W \in \mathbb{R}^{d_y \times d_x}$, $R \in \mathbb{R}^{d_x \times d_z}$, and $u \in \mathbb{R}^{d_u}$. Here, $z(x; u) \in \mathbb{R}^{d_z}$ represents an output of deep residual functions with a parameter vector $u$. No assumption is imposed on the form of $z(x; u)$, and $z(x; u)$ can represent an output of possibly complicated deep residual functions that arise in ResNets. For example, the function $f$ can represent deep preactivation ResNets (He, Zhang, Ren, & Sun, 2016), which are widely used in practice. To simplify theoretical study, Shamir (2018) assumed that every entry of the matrix $R$ is unconstrained (e.g., instead of $R$ representing convolutions). We adopt this assumption based on the previous study (Shamir, 2018).

*4.1.1 Background.* Along with an analysis of approximate critical points, Shamir (2018) proved the following main result, proposition 1, under the assumptions PA1, PA2, and PA3:

PA1: The output dimension $d_y = 1$.
PA2: For any $y$, the function $\ell_y$ is convex and twice differentiable.
PA3: On any bounded subset of the domain of $L$, the function $L_u(W, R)$, its gradient $\nabla L_u(W, R)$, and its Hessian $\nabla^2 L_u(W, R)$ are all Lipschitz continuous in $(W, R)$, where $L_u(W, R) = L(\theta)$ with a fixed $u$.

**Proposition 1** *(Shamir, 2018). Let $f$ be specified by equation 4.1, Let assumptions PA1, PA2, and PA3 hold. Then for any local minimum $\theta$ of L,*

$$L(\theta) \leq \inf_{W \in \mathbb{R}^{d_y \times d_x}} \sum_{i=1}^{m} \lambda_i \ell_{y_i}(Wx_i).$$

Shamir (2018) remarked that it is an open problem whether proposition 1 and another main result in the article can be extended to networks with $d_y > 1$ (multiple output units). Note that Shamir (2018) also provided proposition 1 with an expected loss and an analysis for a simpler decoupled model, $Wx + Vz(x; u)$. For the simpler decoupled model, our theorem 1 immediately concludes that given any $u$, every critical point with respect to $\theta_{-u} = (W, R)$ achieves a global minimum value with respect to $\theta_{-u}$ as $L(\theta_{-u}) = \inf \{\sum_{i=1}^{m} \lambda_i \ell_{y_i}(Wx_i + Rz(x_i; u)) : W \in \mathbb{R}^{d_y \times d_x}, R \in \mathbb{R}^{d_x \times d_z}\}$ ($\leq \inf_{W \in \mathbb{R}^{d_y \times d_x}} \sum_{i=1}^{m} \lambda_i \ell_{y_i}(Wx_i)$). This holds for every critical point $\theta$ since any critical point $\theta$ must be a critical point with respect to $\theta_{-u}$.

**4.2 Result.** The following theorem shows that every differentiable local minimum achieves the global minimum value of $\tilde{L}_\theta^{(\text{ResNet})}$ (the right-hand side in equation 4.2), which is no worse than the upper bound in proposition 1 and is strictly better than the upper bound as long as $z(x_i, u)$ or $\tilde{f}_\theta(x_i; \alpha, \epsilon, S)$ is nonnegligible. Indeed, the global minimum value of $\tilde{L}_\theta^{(\text{ResNet})}$ (the right-hand side in equation 4.2) is no worse than the global minimum value of all models parameterized by the coefficients of the basis $x$ and $z(x; u)$, and further improvement is guaranteed through a nonnegligible $\tilde{f}_\theta(x_i; \alpha, \epsilon, S)$.

**Theorem 3.** *Let $f$ be specified by equation 4.1. Let assumption 1 hold. Assume that $d_y \leq \min\{d_x, d_z\}$. Then for any local minimum $\theta \in (\mathbb{R}^{d_\theta} \setminus \tilde{\Omega})$ of $L$, the* *following holds: there exists $\epsilon_0 > 0$ such that for any $\epsilon \in (0, \epsilon_0)$,*

$$L(\theta) = \inf_{\substack{S \subseteq_{fin} \mathcal{V}[\theta, \epsilon], \\ \alpha \in \mathbb{R}^{d_\theta \times |S|}, \\ \alpha_w \in \mathbb{R}^{d_y \times d_x}, \alpha_r \in \mathbb{R}^{d_y \times d_z}}} \tilde{L}_\theta^{(ResNet)}(\alpha, \alpha_w, \alpha_r, \epsilon, S), \quad (4.2)$$

*where*

$$\tilde{L}_\theta^{(ResNet)}(\alpha, \alpha_w, \alpha_r, \epsilon, S) = \sum_{i=1}^m \lambda_i \ell_{y_i}(\alpha_w x_i + \alpha_r z(x_i; u) + \tilde{f}_\theta(x_i; \alpha, \epsilon, S)).$$

Theorem 3 also successfully solved the first part of the open problem in the literature (Shamir, 2018) by discarding the assumption of $d_y = 1$. From the geometric view, theorem 3 states that the span $\tilde{T}_{f_X(\theta)}$ of the set of the vectors in the tangent spaces $\{T_{f_X(\theta+\epsilon v)} : v \in \mathcal{V}[\theta, \epsilon]\}$ contains the output of the best basis model with the linear feature $x$ and the learned nonlinear feature $z(x_i; u)$. Similar to the examples in Figures 3 and 4, $\tilde{T}_{f_X(\theta)} \neq T_{f(\theta)}$ and the output of the best basis model with these features is contained in $\tilde{T}_{f_X(\theta)}$ but not in $T_{f(\theta)}$.

Unlike the recent study on ResNets (Kawaguchi & Bengio, 2019), our theorem 3 predicts the value of $L$ through the global minimum value of a large search space (i.e., the domain of $\tilde{L}_\theta^{(ResNet)}$) and is proven as a consequence of our general theory (i.e., theorem 2) with a significantly different proof idea (see section 4.3) and with the novel geometric insight.

*4.2.1 Example: Deep Nonlinear Networks with Locally Induced Partial Linear Structures.* We specify $f$ to represent fully connected feedforward networks with arbitrary nonlinearity $\sigma$ and arbitrary depth $H$ as follows:

$$f(x; \theta) = W^{(H+1)} h^{(H)}(x; \theta), \quad (4.3)$$

where

$$h^{(l)}(x; \theta) = \sigma^{(l)}(W^{(l)} h^{(l-1)}(x; \theta)),$$

for all $l \in \{1, \ldots, H\}$ with $h^{(0)}(x; \theta) = x$. Here, $\theta = \text{vec}([W^{(l)}]_{l=1}^{H+1}) \in \mathbb{R}^{d_\theta}$ with $W^{(l)} \in \mathbb{R}^{d_l \times d_{l-1}}$, $d_{H+1} = d_y$, and $d_0 = d_x$. In addition, $\sigma^{(l)} : \mathbb{R}^{d_l} \to \mathbb{R}^{d_l}$ represents an arbitrary nonlinear activation function per layer $l$ and is allowed to differ among different layers.

*4.2.2 Background.* Given the difficulty of theoretically understanding deep neural networks, Goodfellow, Bengio, and Courville (2016) noted that theoretically studying simplified networks (i.e., deep linear networks) is worthwhile. For example, Saxe, McClelland, and Ganguli (2014) empirically showed that deep linear networks may exhibit several properties analogous to those of deep nonlinear networks. Accordingly, the theoretical study of deep linear neural networks has become an active area of research (Kawaguchi, 2016; Hardt & Ma, 2017; Arora, Cohen, Golowich, & Hu, 2018; Arora, Cohen, & Hazan, 2018; Bartlett, Helmbold, & Long, 2019; Du & Hu, 2019).

Along this line, Laurent and Brecht (2018) recently proved the following main result, proposition 2, under the assumptions PA4, PA5, and PA6:

PA4: Every activation function is identity as $\sigma^{(l)}(q) = q$ for every $l \in \{1, \ldots, H\}$ (i.e., deep linear networks).
PA5: For any $y$, the function $\ell_y$ is convex and differentiable.
PA6: The thinnest layer is either the input layer or the output layer as $\min\{d_x, d_y\} \leq \min\{d_1, \ldots, d_H\}$.

**Proposition 2** *(Laurent & Brecht, 2018). Let $f$ be specified by equation 4.3. Let assumptions PA4, PA5, and PA6 hold. Then every local minimum $\theta$ of L is a global minimum.*

*4.2.3 Result.* Instead of studying deep linear networks, we now consider a partial linear structure locally induced by a parameter vector with nonlinear activation functions. This relaxes the linearity assumption and extends our understanding of deep linear networks to deep nonlinear networks.

Intuitively, $\mathcal{J}_{n,t}[\theta]$ is a set of partial linear structures locally induced by a vector $\theta$, which is now formally defined as follows. Given a $\theta \in \mathbb{R}^{d_\theta}$, let $\mathcal{J}_{n,t}[\theta]$ be a set of all sets $J = \{J^{(t+1)}, \ldots, J^{(H+1)}\}$ such that each set $J = \{J^{(t+1)}, \ldots, J^{(H+1)}\} \in \mathcal{J}_{n,t}[\theta]$ satisfies the following conditions: there exists $\epsilon > 0$ such that for all $l \in \{t+1, t+2, \ldots, H+1\}$,

1. $J^{(l)} \subseteq \{1, \ldots, d_l\}$ with $|J^{(l)}| \geq n$.
2. $h^{(l)}(x_i, \theta')_k = (W^{(l)} h^{(l-1)}(x_i, \theta'))_k$ for all $(k, \theta', i) \in J^{(l)} \times B(\theta, \epsilon) \times \{1, \ldots, m\}$.
3. $W^{(l+1)}_{i,j} = 0$ for all $(i, j) \in (\{1, \ldots, d_{l+1}\} \setminus J^{(l+1)}) \times J^{(l)}$ if $l \leq H - 1$.

Let $\Theta_{n,t}$ be the set of all parameter vectors $\theta$ such that $\mathcal{J}_{n,t}[\theta]$ is nonempty. As the definition reveals, a neural network with a $\theta \in \Theta_{d_y,t}$ can be a standard deep nonlinear neural network (with no linear units).

**Theorem 4.** *Let $f$ be specified by equation 4.3. Let assumption 1 hold. Then for any $t \in \{1, \ldots, H\}$, at every local minimum $\theta \in (\Theta_{d_y,t} \setminus \tilde{\Omega})$ of L, the following holds. There exists $\epsilon_0 > 0$ such that for any $\epsilon \in (0, \epsilon_0)$,*

$$L(\theta) = \inf_{\substack{S \subseteq_{fin} \mathcal{V}[\theta, \epsilon], \\ \alpha \in \mathbb{R}^{d_\theta \times |S|}, \alpha_h \in \mathbb{R}^{d_t}}} \tilde{L}^{(ff)}_{\theta, t}(\alpha, \alpha_h, \epsilon, S),$$

*where*

$$\tilde{L}_{\theta,t}^{(ff)}(\alpha, \alpha_h, \epsilon, S) = \sum_{i=1}^{m} \lambda_i \ell_{y_i} \left( \sum_{l=t}^{H} \alpha_h^{(l+1)} h^{(l)}(x_i; u) + \tilde{f}_\theta(x_i; \alpha, \epsilon, S) \right),$$

with $\alpha_h = vec([\alpha_h^{(l+1)}]_{l=t}^{H}) \in \mathbb{R}^{d_t}$, $\alpha_h^{(l+1)} \in \mathbb{R}^{d_y \times d_l}$ and $d_t = d_y \sum_{l=t}^{H} d_l$.

Theorem 4 is a special case of theorem 2. A special case of theorem 4 then results in one of the main results in the literature regarding deep linear neural networks, that is, every local minimum is a global minimum. Consider any deep linear network with $d_y \leq \min\{d_1, \ldots, d_H\}$. Then every local minimum $\theta$ is in $\Theta_{d_y,0} \setminus \tilde{\Omega} = \Theta_{d_y,0}$. Hence, theorem 4 is reduced to the statement that for any local minimum, $L(\theta) = \inf_{\alpha_h \in \mathbb{R}^{d_t}} \sum_{i=1}^{m} \lambda_i \ell_{y_i}(\sum_{l=0}^{H} \alpha_h^{(l+1)} h^{(l)}(x_i; u)) = \inf_{\alpha_x \in \mathbb{R}^{d_x}} \sum_{i=1}^{m} \lambda_i \ell_{y_i}(\alpha_x x_i)$, which is the global minimum value. Thus, every local minimum is a global minimum for any deep linear neural network with $d_y \leq \min\{d_1, \ldots, d_H\}$. Therefore, theorem 4 successfully generalizes the recent previous result in the literature (proposition 2) for a common scenario of $d_y \leq d_x$.

Beyond deep linear networks, theorem 4 illustrates both the benefit of the locally induced structure and the overparameterization for deep nonlinear networks. In the first term, $\sum_{l=t}^{H} \alpha_h^{(l+1)} h^{(l)}(x_i; u)$, in $L_{\theta,t}^{(ff)}$, we benefit by decreasing $t$ (a more locally induced structure) and increasing the width of the $l$th layer for any $l \geq t$ (overparameterization). The second term, $\tilde{f}_\theta(x_i; \alpha, \epsilon, S)$ in $L_{\theta,t}^{(ff)}$, is the general term that is always present from theorem 2, where we benefit from increasing $d_\theta$ because $\alpha \in \mathbb{R}^{d_\theta \times |S|}$.

From the geometric view, theorem 4 captures the intuition that the span $\tilde{T}_{f_X(\theta)}$ of the set of the vectors in the tangent spaces $\{T_{f_X(\theta+\epsilon v)} : v \in \mathcal{V}[\theta, \epsilon]\}$ contains the best basis model with the linear feature for deep linear networks, as well as the best basis models with more nonlinear features as more local structures arise. Similar to the examples in Figures 3 and 4, $\tilde{T}_{f_X(\theta)} \neq T_{f(\theta)}$ and the output of the best basis models with those features are contained in $\tilde{T}_{f_X(\theta)}$ but not in $T_{f(\theta)}$.

A similar local structure was recently considered in Kawaguchi, Huang, and Kaelbling (2019). However, both the problem settings and the obtained results largely differ from those in Kawaguchi et al. (2019). Furthermore, theorem 4 is proven as a consequence of our general theory (theorem 2), and accordingly, the proofs largely differ from each other as well. Theorem 4 also differs from recent results on the gradient decent algorithm for deep linear networks (Arora, Cohen, Golowich, & Hu, 2018; Arora, Cohen, & Hazan, 2018; Bartlett et al., 2019; Du & Hu, 2019), since we analyze the loss surface instead of a specific algorithm and theorem 4 applies to deep nonlinear networks as well.

**4.3 Proof Idea in Applications of Theorem 2.** Theorems 3 and 4 are simple consequences of theorem 2, and their proof is illustrative as a means of using theorem 2 in future studies with different additional assumptions. The high-level idea behind the proofs in the applications of theorem 2 is captured in the geometric view of theorem 2 (see Figures 3 and 4). That is, given a desired guarantee, we check whether the space $\tilde{T}_{f_X(\theta)}$ is expressive enough to contain the output of the desired model corresponding to the desired guarantee.

To simplify the use of theorem 2, we provide the following lemma. This lemma states that the expressivity of the model $\tilde{f}_\theta(x; \alpha, \epsilon, S)$ with respect to $(\alpha, S)$ is the same as that of $\tilde{f}_\theta(x; \alpha, \epsilon, S) + \tilde{f}_\theta(x; \alpha', \epsilon, S')$ with respect to $(\alpha, \alpha', S, S')$. As shown in its proof, this is essentially because $\tilde{f}_\theta$ is linear in $\alpha$, and a union of two sets $S \subseteq_{\text{fin}} \mathcal{V}[\theta, \epsilon]$ and $S' \subseteq_{\text{fin}} \mathcal{V}[\theta, \epsilon]$ remains a finite subset of $\mathcal{V}[\theta, \epsilon]$.

**Lemma 1.** *For any $\theta$, any $\epsilon \geq 0$, any $S' \subseteq_{fin} \mathcal{V}[\theta, \epsilon]$, and any $x$, it holds that $\{\tilde{f}_\theta(x; \alpha, \epsilon, S) : \alpha \in \mathbb{R}^{d_\theta \times |S|}, S \subseteq_{fin} \mathcal{V}[\theta, \epsilon]\} = \{\tilde{f}_\theta(x; \alpha, \epsilon, S) + \tilde{f}_\theta(x; \alpha', \epsilon, S') : \alpha \in \mathbb{R}^{d_\theta \times |S|}, \alpha' \in \mathbb{R}^{d_\theta \times |S'|}, S \subseteq_{fin} \mathcal{V}[\theta, \epsilon]\}$.*

Based on theorem 2 and lemma 1, the proofs of theorems 3 and 4 are reduced to a simple search for finding $S' \subseteq_{\text{fin}} \mathcal{V}[\theta, \epsilon]$ such that the expressivity of $\tilde{f}_\theta(x_i; \alpha', \epsilon, S')$ with respect to $\alpha'$ is no worse than the expressivity of $\alpha_w x_i + \alpha_r z(x_i; u)$ with respect to $(\alpha_w, \alpha_r)$ (see theorem 3) and that of $\sum_{l=t}^{H} \alpha_h^{(l+1)} h^{(l)}(x_i; u)$ with respect to $\alpha_h^{(l+1)}$ (see theorem 4). In other words, $\{\tilde{f}_\theta(x_i; \alpha', \epsilon, S') : \alpha' \in \mathbb{R}^{d_\theta \times |S'|}\} \supseteq \{\alpha_w x_i + \alpha_r z(x_i; u) : \alpha_w \in \mathbb{R}^{d_y \times d_x}, \alpha_r \in \mathbb{R}^{d_y \times d_z}\}$ (see theorem 3) and $\{\tilde{f}_\theta(x_i; \alpha', \epsilon, S') : \alpha' \in \mathbb{R}^{d_\theta \times |S'|}\} \supseteq \{\sum_{l=t}^{H} \alpha_h^{(l+1)} h^{(l)}(x_i; u) : \alpha_h \in \mathbb{R}^{d_t}\}$ (see theorem 4). Only with this search for $S'$, theorem 2 together with lemma 1 implies the desired statements for theorems 3 and 4 (see sections A.4 and A.5 in the appendix for further details). Thus, theorem 2 also enables simple proofs.

## 5 Conclusion

This study provided a general theory for nonconvex machine learning and demonstrated its power by proving new competitive theoretical results with it. In general, the proposed theory provides a mathematical tool to study the effects of hypothesis classes $f$, methods, and assumptions through the lens of the global optima of the perturbable gradient basis model class.

In convex machine learning with a model output $f(x; \theta) = \theta^\top x$ with a (nonlinear) feature output $x = \phi(x^{(\text{raw})})$, achieving a critical point ensures the global optimality in the span of the fixed basis $x = \phi(x^{(\text{raw})})$. In nonconvex machine learning, we have shown that achieving a critical point ensures the global optimality in the span of the gradient basis $\partial f_x(\theta)$, which coincides with the fixed basis $x = \phi(x^{(\text{raw})})$ in the case of the convex machine learning. Thus, whether convex or nonconvex, achieving a critical point ensures the global optimality in the span of some basis, which might be arbitrarily bad (or good) depending on the choice of the handcrafted basis $\phi(x^{(\text{raw})}) = \partial f_x(\theta)$ (for the convex case) or the induced basis $\partial f_x(\theta)$ (for the nonconvex case). Therefore, in terms of the loss values at critical points, nonconvex machine learning is theoretically as justified as the convex one, except in the case when a preference is given to $\phi(x^{(\text{raw})})$ over $\partial f_x(\theta)$ (both of which can be arbitrarily bad or good). The same statement holds for local minima and perturbable gradient basis.

## Appendix: Proofs of Theoretical Results

In this appendix, we provide complete proofs of the theoretical results.

**A.1 Proof of Theorem 1.** The proof of theorem 1 combines lemma 2 with assumptions 1 and 2 by taking advantage of the structure of the objective function $L$. Although lemma 2 is rather weak and assumptions 1 and 2 are mild (in the sense that they usually hold in practice), a right combination of these with the structure of $L$ can prove the desired statement.

**Lemma 2.** *Assume that for any $i \in \{1, \ldots, m\}$, the function $\ell_{y_i} : q \mapsto \ell(q, y_i)$ is differentiable. Then for any critical point $\theta \in (\mathbb{R}^{d_\theta} \setminus \Omega)$ of $L$, the following holds: for any $k \in \{1, \ldots, d_\theta\}$,*

$$\sum_{i=1}^{m} \lambda_i \partial \ell_{y_i}(f_{x_i}(\theta)) \partial_k f_{x_i}(\theta) = 0.$$

**Proof of Lemma 2.** Let $\theta$ be an arbitrary critical point $\theta \in (\mathbb{R}^{d_\theta} \setminus \Omega)$ of $L$. Since $\ell_{y_i} : \mathbb{R}^{d_y} \to \mathbb{R}$ is assumed to be differentiable and $f_{x_i} \in \mathbb{R}^{d_y}$ is differentiable at the given $\theta$, the composition $(\ell_{y_i} \circ f_{x_i})$ is also differentiable, and $\partial_k(\ell_{y_i} \circ f_{x_i}) = \partial \ell_{y_i}(f_{x_i}(\theta)) \partial_k f_{x_i}(\theta)$. In addition, $L$ is differentiable because a sum of differentiable functions is differentiable. Therefore, for any critical point $\theta$ of $L$, we have that $\partial L(\theta) = 0$, and, hence, $\partial_k L(\theta) = \sum_{i=1}^{m} \lambda_i \partial \ell_{y_i}(f_{x_i}(\theta)) \partial_k f_{x_i}(\theta) = 0$, for any $k \in \{1, \ldots, d_\theta\}$, from linearity of differentiation operation. □

**Proof of Theorem 1.** Let $\theta \in (\mathbb{R}^{d_\theta} \setminus \Omega)$ be an arbitrary critical point of $L$. From assumption 2, there exists a function $g$ such that $f_{x_i}(\theta) = \sum_{k=1}^{d_\theta} g(\theta)_k \partial_k f_{x_i}(\theta)$ for all $i \in \{1, \ldots, m\}$. Then, for any $\alpha \in \mathbb{R}^{d_\theta}$,

$$L_\theta(\alpha) \geq \sum_{i=1}^{m} \lambda_i \ell_{y_i}(f_{x_i}(\theta)) + \lambda_i \partial \ell_{y_i}(f_{x_i}(\theta))(f_\theta(x_i; \alpha) - f(x_i; \theta))$$

$$= \sum_{i=1}^{m} \lambda_i \ell_{y_i}(f_{x_i}(\theta)) + \sum_{k=1}^{d_\theta} \alpha_k \underbrace{\sum_{i=1}^{m} \lambda_i \partial \ell_{y_i}(f_{x_i}(\theta)) \partial_k f_{x_i}(\theta)}_{=0 \text{ from Lemma 2}}$$

$$- \sum_{i=1}^{m} \lambda_i \partial \ell_{y_i}(f_{x_i}(\theta)) f(x_i; \theta)$$

$$= \sum_{i=1}^{m} \lambda_i \ell_{y_i}(f_{x_i}(\theta)) - \sum_{k=1}^{d_\theta} g(\theta)_k \underbrace{\sum_{i=1}^{m} \lambda_i \partial \ell_{y_i}(f_{x_i}(\theta)) \partial_k f_{x_i}(\theta)}_{=0 \text{ from Lemma 2}},$$

$$= L(\theta),$$

where the first line follows from assumption 1 (differentiable and convex $\ell_{y_i}$), the second line follows from linearity of summation, and the third line follows from assumption 2. Thus, on the one hand, we have that $L(\theta) \leq \inf_{\alpha \in \mathbb{R}^{d_\theta}} L_\theta(\alpha)$. On the other hand, since $f(x_i; \theta) = \sum_{k=1}^{d_\theta} g(\theta)_k \partial_k f_{x_i}(\theta) \in \{f_\theta(x_i; \alpha) = \sum_{k=1}^{d_\theta} \alpha_k \partial_k f_{x_i}(\theta) : \alpha \in \mathbb{R}^{d_\theta}\}$, we have that $L(\theta) \geq \inf_{\alpha \in \mathbb{R}^{d_\theta}} L_\theta(\alpha)$. Combining these yields the desired statement of $L(\theta) = \inf_{\alpha \in \mathbb{R}^{d_\theta}} L_\theta(\alpha)$. □

**A.2 Proof of Theorem 2.** The proof of theorem 2 uses lemma 3, the structure of the objective function $L$, and assumptions 1 and 2.

**Lemma 3.** *Assume that for any $i \in \{1, \ldots, m\}$, the function $\ell_{y_i} : q \mapsto \ell(q, y_i)$ is differentiable. Then for any local minimum $\theta \in (\mathbb{R}^{d_\theta} \setminus \tilde{\Omega})$ of $L$, the following holds: there exists $\epsilon_0 > 0$ such that for any $\epsilon \in [0, \epsilon_0)$, any $v \in \mathcal{V}[\theta, \epsilon]$, and any $k \in \{1, \ldots, d_\theta\}$,*

$$\sum_{i=1}^{m} \lambda_i \partial \ell_{y_i}(f_{x_i}(\theta)) \partial_k f_{x_i}(\theta + \epsilon v) = 0.$$

**Proof of Lemma 3.** Let $\theta \in (\mathbb{R}^{d_\theta} \setminus \tilde{\Omega})$ be an arbitrary local minimum of $L$. Since $\theta$ is a local minimum of $L$, by the definition of a local minimum, there exists $\epsilon_1 > 0$ such that $L(\theta) \leq L(\theta')$ for all $\theta' \in B(\theta, \epsilon_1)$. Then for any $\epsilon \in [0, \epsilon_1/2)$ and any $v \in \mathcal{V}[\theta, \epsilon]$, the vector $(\theta + \epsilon v)$ is also a local minimum because

$$L(\theta + \epsilon v) = L(\theta) \leq L(\theta')$$

for all $\theta' \in B(\theta + \epsilon v, \epsilon_1/2) \subseteq B(\theta, \epsilon_1)$ (the inclusion follows from the triangle inequality), which satisfies the definition of a local minimum for $(\theta + \epsilon v)$.

Since $\theta \in (\mathbb{R}^{d_\theta} \setminus \tilde{\Omega})$, there exists $\epsilon_2 > 0$ such that $f_{x_1}, \ldots, f_{x_m}$ are differentiable in $B(\theta, \epsilon_2)$. Since $\ell_{y_i} : \mathbb{R}^{d_y} \to \mathbb{R}$ is assumed to be differentiable and $f_{x_i} \in \mathbb{R}^{d_y}$ is differentiable in $B(\theta, \epsilon_2)$, the composition $(\ell_{y_i} \circ f_{x_i})$ is also differentiable, and $\partial_k(\ell_{y_i} \circ f_{x_i}) = \partial \ell_{y_i}(f_{x_i}(\theta))\partial_k f_{x_i}(\theta)$ in $B(\theta, \epsilon_2)$. In addition, $L$ is differentiable in $B(\theta, \epsilon_2)$ because a sum of differentiable functions is differentiable.

Therefore, with $\epsilon_0 = \min(\epsilon_1/2, \epsilon_2)$, we have that for any $\epsilon \in [0, \epsilon_0)$ and any $v \in \mathcal{V}[\theta, \epsilon]$, the vector $(\theta + \epsilon v)$ is a differentiable local minimum, and hence the first-order necessary condition of differentiable local minima implies that

$$\partial_k L(\theta + \epsilon v) = \sum_{i=1}^m \lambda_i \partial \ell_{y_i}(f_{x_i}(\theta))\partial_k f_{x_i}(\theta + \epsilon v) = 0,$$

for any $k \in \{1, \ldots, d_\theta\}$, where we used the fact that $f_{x_i}(\theta) = f_{x_i}(\theta + \epsilon v)$ for any $v \in \mathcal{V}[\theta, \epsilon]$. □

**Proof of Theorem 2.** Let $\theta \in (\mathbb{R}^{d_\theta} \setminus \tilde{\Omega})$ be an arbitrary local minimum of $L$. Since $(\mathbb{R}^{d_\theta} \setminus \tilde{\Omega}) \subseteq (\mathbb{R}^{d_\theta} \setminus \Omega)$, from assumption 2, there exists a function $g$ such that $f_{x_i}(\theta) = \sum_{k=1}^{d_\theta} g(\theta)_k \partial_k f_{x_i}(\theta)$ for all $i \in \{1, \ldots, m\}$. Then from lemma 3, there exists $\epsilon_0 > 0$ such that for any $\epsilon \in [0, \epsilon_0)$, any $S \subseteq_{\text{fin}} \mathcal{V}[\theta, \epsilon]$ and any $\alpha \in \mathbb{R}^{d_\theta \times |S|}$,

$$\tilde{L}_\theta(\alpha, \epsilon, S) \geq \sum_{i=1}^m \lambda_i \ell_{y_i}(f_{x_i}(\theta)) + \lambda_i \partial \ell_{y_i}(f_{x_i}(\theta))(\tilde{f}_\theta(x_i; \alpha, \epsilon, S) - f(x_i; \theta))$$

$$= \sum_{i=1}^m \lambda_i \ell_{y_i}(f_{x_i}(\theta)) + \sum_{k=1}^{d_\theta} \sum_{j=1}^{|S|} \alpha_{k,j} \underbrace{\sum_{i=1}^m \lambda_i \partial \ell_{y_i}(f_{x_i}(\theta))\partial_k f_{x_i}(\theta + \epsilon S_j)}_{=0 \text{ from Lemma 3}}$$

$$- \sum_{i=1}^m \lambda_i \partial \ell_{y_i}(f_{x_i}(\theta)) f(x_i; \theta)$$

$$= \sum_{i=1}^m \lambda_i \ell_{y_i}(f_{x_i}(\theta)) - \sum_{k=1}^{d_\theta} g(\theta)_k \underbrace{\sum_{i=1}^m \lambda_i \partial \ell_{y_i}(f_{x_i}(\theta))\partial_k f_{x_i}(\theta)}_{=0 \text{ from Lemma 3}},$$

$$= L(\theta),$$

where the first line follows from assumption 1 (differentiable and convex $\ell_{y_i}$), the second line follows from linearity of summation and the definition of $\tilde{f}_\theta(x_i; \alpha, \epsilon, S)$, and the third line follows from assumption 2.

Thus, on the one hand, there exists $\epsilon_0 > 0$ such that for any $\epsilon \in [0, \epsilon_0)$, $L(\theta) \leq \inf\{\tilde{L}_\theta(\alpha, \epsilon, S) : S \subseteq_{\text{fin}} \mathcal{V}[\theta, \epsilon], \alpha \in \mathbb{R}^{d_\theta \times |S|}\}$. On the other hand, since $f(x_i; \theta) = \sum_{k=1}^{d_\theta} g(\theta)_k \partial_k f_{x_i}(\theta) \in \{\tilde{f}_\theta(x_i; \alpha, \epsilon, S) : \alpha \in \mathbb{R}^{d_\theta}, S = \emptyset\}$, we have that $L(\theta) \leq \inf\{\tilde{L}_\theta(\alpha, \epsilon, S) : S \subseteq_{\text{fin}} \mathcal{V}[\theta, \epsilon], \alpha \in \mathbb{R}^{d_\theta \times |S|}\}$. Combining these yields the desired statement. □

**A.3 Proof of Lemma 1.** As shown in the proof of lemma 1, lemma 1 is a simple consequence of the following facts: $\tilde{f}_\theta$ is linear in $\alpha$ and a union of two sets $S \subseteq_{\text{fin}} \mathcal{V}[\theta, \epsilon]$ and $S' \subseteq_{\text{fin}} \mathcal{V}[\theta, \epsilon]$ is still a finite subset of $\mathcal{V}[\theta, \epsilon]$.

**Proof of Lemma 1.** Let $S' \subseteq_{\text{fin}} \mathcal{V}[\theta, \epsilon]$ be fixed. Then,

$$\{\tilde{f}_\theta(x; \alpha, \epsilon, S) : \alpha \in \mathbb{R}^{d_\theta \times |S|}, S \subseteq_{\text{fin}} \mathcal{V}[\theta, \epsilon]\}$$
$$= \{\tilde{f}_\theta(x; \alpha, \epsilon, S \cup S') : \alpha \in \mathbb{R}^{d_\theta \times |S \cup S'|}, S \subseteq_{\text{fin}} \mathcal{V}[\theta, \epsilon]\}$$
$$= \{\tilde{f}_\theta(x; \alpha, \epsilon, S \setminus S') + \tilde{f}_\theta(x; \alpha', \epsilon, S') : \alpha \in \mathbb{R}^{d_\theta \times |S \setminus S'|}, \alpha' \in \mathbb{R}^{d_\theta \times |S'|},$$
$$\quad S \subseteq_{\text{fin}} \mathcal{V}[\theta, \epsilon]\}$$
$$= \{\tilde{f}_\theta(x; \alpha, \epsilon, S \cup S') + f_\theta(x; \alpha', \epsilon, S') : \alpha \in \mathbb{R}^{d_\theta \times |S \cup S'|}, \alpha' \in \mathbb{R}^{d_\theta \times |S'|},$$
$$\quad S \subseteq_{\text{fin}} \mathcal{V}[\theta, \epsilon]\}$$
$$= \{\tilde{f}_\theta(x; \alpha, \epsilon, S) + f_\theta(x; \alpha', \epsilon, S') : \alpha \in \mathbb{R}^{d_\theta \times |S|}, \alpha' \in \mathbb{R}^{d_\theta \times |S'|},$$
$$\quad S \subseteq_{\text{fin}} \mathcal{V}[\theta, \epsilon]\},$$

where the second line follows from the facts that a finite union of finite sets is finite and hence $S \cup S' \subseteq_{\text{fin}} \mathcal{V}[\theta, \epsilon]$ (i.e., the set in the first line is a superset of $\supseteq$, the set in the second line), and that $\alpha \in \mathbb{R}^{d_\theta \times |S \cup S'|}$ can vanish the extra terms due to $S'$ in $\tilde{f}_\theta(x; \alpha, \epsilon, S \cup S')$ (i.e., the set in the first line is a subset of, $\subseteq$, the set in the second line). The last line follows from the same facts. The third line follows from the definition of $\tilde{f}_\theta(x; \alpha, \epsilon, S)$. The fourth line follows from the following equality due to the linearity of $\tilde{f}_\theta$ in $\alpha$:

$$\{\tilde{f}_\theta(x; \alpha', \epsilon, S') : \alpha' \in \mathbb{R}^{d_\theta \times |S'|}\}$$
$$= \left\{\sum_{k=1}^{d_\theta} \sum_{j=1}^{|S|} (\alpha'_{k,j} + \bar{\alpha}'_{k,j}) \partial_k f_x(\theta + \epsilon S'_j) : \alpha' \in \mathbb{R}^{d_\theta \times |S'|}, \bar{\alpha}' \in \mathbb{R}^{d_\theta \times |S'|}\right\}$$
$$= \{\tilde{f}_\theta(x; \alpha', \epsilon, S') + \tilde{f}_\theta(x; \bar{\alpha}', \epsilon, S') : \alpha' \in \mathbb{R}^{d_\theta \times |S'|}, \bar{\alpha}' \in \mathbb{R}^{d_\theta \times |S'|}\}.$$

□

**A.4 Proof of Theorem 3.** As shown in the proof of theorem 3, thanks to theorem 2 and lemma 1, the remaining task to prove theorem 3 is to find a set $S' \subseteq_{\text{fin}} \mathcal{V}[\theta, \epsilon]$ such that $\{\tilde{f}_\theta(x_i; \alpha', \epsilon, S') : \alpha' \in \mathbb{R}^{d_\theta \times |S'|}\} \supseteq \{\alpha_w x_i + \alpha_r z(x_i; u) : \alpha_w \in \mathbb{R}^{d_y \times d_x}, \alpha_r \in \mathbb{R}^{d_y \times d_z}\}$. Let Null($M$) be the null space of a matrix $M$.

**Proof of Theorem 3.** Let $\theta \in (\mathbb{R}^{d_\theta} \setminus \tilde{\Omega})$ be an arbitrary local minimum of $L$. Since $f$ is specified by equation 4.1, and hence $f(x; \theta) = (\partial_{\text{vec}(W)} f(x; \theta)) \text{vec}(W)$, assumption 2 is satisfied. Thus, from theorem 2, there exists $\epsilon_0 > 0$ such that for any $\epsilon \in [0, \epsilon_0)$,

$$L(\theta) = \inf_{S \subseteq_{\text{fin}} \mathcal{V}[\theta, \epsilon], \alpha \in \mathbb{R}^{d_\theta \times |S|}} \sum_{i=1}^{m} \lambda_i \ell(\tilde{f}_\theta(x_i; \alpha, \epsilon, S), y_i),$$

where

$$\tilde{f}_\theta(x_i; \alpha, \epsilon, S) = \sum_{j=1}^{|S|} \alpha_{w,j}(x_i + (R + \epsilon v_{r,j})z_{i,j}) + (W + \epsilon v_{w,j})\alpha_{r,j} z_{i,j}$$
$$+ (\partial_u f_{x_i}(\theta + \epsilon S_j))\alpha_{u,j},$$

with $\alpha = [\alpha_{\cdot 1}, \dots, \alpha_{\cdot |S|}] \in \mathbb{R}^{d_\theta \times |S|}$, $\alpha_{\cdot j} = \text{vec}([\alpha_{w,j}, \alpha_{r,j}, \alpha_{u,j}]) \in \mathbb{R}^{d_\theta}$, $S_j = \text{vec}([v_{w,j}, v_{r,j}, v_{u,j}]) \in \mathbb{R}^{d_\theta}$, and $z_{i,j} = z(x_i, u + \epsilon v_{u,j})$ for all $j \in \{1, \dots, |S|\}$. Here, $\alpha_{w,j}, v_{w,j} \in \mathbb{R}^{d_y \times d_x}$, $\alpha_{r,j}, v_{r,j} \in \mathbb{R}^{d_x \times d_z}$, and $\alpha_{u,j}, v_{u,j} \in \mathbb{R}^{d_u}$. Let $\epsilon \in (0, \epsilon_0)$ be fixed.

Consider the case of rank($W$) $\geq d_y$. Define $\bar{S}$ such that $|\bar{S}| = 1$ and $\bar{S}_1 = 0 \in \mathbb{R}^{d_\theta}$, which is in $\mathcal{V}[\theta, \epsilon]$. Then by setting $\alpha_{u,1} = 0$ and rewriting $\alpha_{r,1}$ such that $W \alpha_{r,1} = \alpha_{r,1}^{(1)} - \alpha_{w,1} R$ with an arbitrary matrix $\alpha_{r,1} \in \mathbb{R}^{d_y \times d_z}$ (this is possible since rank($W$) $\geq d_y$), we have that

$$\{\tilde{f}_\theta(x_i; \alpha, \epsilon, \bar{S}) : \alpha \in \mathbb{R}^{d_\theta \times |\bar{S}|}\}$$
$$\supseteq \{\alpha_{w,1} x_i + \alpha_{r,1}^{(1)} z_{i,1} : \alpha_{w,1} \in \mathbb{R}^{d_y \times d_x}, \alpha_{r,1}^{(1)} \in \mathbb{R}^{d_y \times d_z}\}.$$

Consider the case of rank($W$) $< d_y$. Since $W \in \mathbb{R}^{d_y \times d_x}$ and rank($W$) $< d_y \leq \min(d_x, d_z) \leq d_x$, we have that Null($W$) $\neq \{0\}$, and there exists a vector $a \in \mathbb{R}^{d_x}$ such that $a \in$ Null($W$) and $\|a\|_2 = 1$. Let $a$ be such a vector. Define $\bar{S}'$ as follows: $|\bar{S}'| = d_y d_z + 1$, $\bar{S}'_1 = 0 \in \mathbb{R}^{d_\theta}$, and set $\bar{S}'_j$ for all $j \in \{2, \dots, d_y d_z + 1\}$ such that $v_{w,j} = 0$, $v_{u,j} = 0$, and $v_{r,j} = ab_j^\top$ where $b_j \in \mathbb{R}^{d_z}$ is an arbitrary column vector with $\|b_j\|_2 \leq 1$. Then $\bar{S}'_j \in \mathcal{V}[\theta, \epsilon]$ for all $j \in \{1, \dots, d_y d_z + 1\}$. By setting $\alpha_{r,j} = 0$ and $\alpha_{u,j} = 0$ for all $j \in \{1, \dots, d_y d_z + 1\}$ and by rewriting $\alpha_{w,1} = \alpha_{w,1}^{(1)} - \sum_{j=2}^{d_y d_z + 1} \alpha_{w,j}$ and $\alpha_{w,j} = \frac{1}{\epsilon} q_j a^T$ for all $j \in \{2, \dots, d_y d_z + 1\}$ with an arbitrary vector $q_j \in \mathbb{R}^{d_y}$ (this is possible since $\epsilon > 0$ is fixed first and $\alpha_{w,j}$ is arbitrary), we have that

$$\{\tilde{f}_\theta(x_i; \alpha, \epsilon, \bar{S}') : \alpha \in \mathbb{R}^{d_\theta \times |\bar{S}'|}\}$$

$$\supseteq \left\{ \alpha_{w,1}^{(1)} x_i + \left( \alpha_{w,1}^{(1)} R + \sum_{j=2}^{d_y d_z + 1} q_j b_j^\top \right) z_{i,1} : q_j \in \mathbb{R}^{d_y}, b_j \in \mathbb{R}^{d_z} \right\}.$$

Since $q_j \in \mathbb{R}^{d_y}$ and $b_j \in \mathbb{R}^{d_z}$ are arbitrary, we can rewrite $\sum_{j=2}^{d_y d_z + 1} q_j b_j = \alpha_{w,1}^{(2)} - \alpha_{w,1}^{(1)} R$ with an arbitrary matrix $\alpha_{w,1}^{(2)} \in \mathbb{R}^{d_y \times d_z}$, yielding

$$\{\tilde{f}_\theta(x_i; \alpha, \epsilon, \bar{S}') : \alpha \in \mathbb{R}^{d_\theta \times |\bar{S}'|}\}$$

$$\supseteq \{\alpha_{w,1}^{(1)} x_i + \alpha_{w,1}^{(2)} z_{i,1} : \alpha_{w,1}^{(1)} \in \mathbb{R}^{d_y \times d_x}, \alpha_{w,1}^{(2)} \in \mathbb{R}^{d_y \times d_z}\}.$$

By summarizing above, in both cases of rank($W$), there exists a set $S' \subseteq_{\text{fin}} \mathcal{V}[\theta, \epsilon]$ such that

$$\{\tilde{f}_\theta(x_i; \alpha, \epsilon, S) : \alpha \in \mathbb{R}^{d_\theta \times |S|}, S \subseteq_{\text{fin}} \mathcal{V}[\theta, \epsilon]\}$$

$$= \{\tilde{f}_\theta(x_i; \alpha, \epsilon, S) + \tilde{f}_\theta(x_i; \alpha', \epsilon, S')$$

$$: \alpha \in \mathbb{R}^{d_\theta \times |S|}, \alpha' \in \mathbb{R}^{d_\theta \times |S'|}, S \subseteq_{\text{fin}} \mathcal{V}[\theta, \epsilon]\}$$

$$\supseteq \{\tilde{f}_\theta(x_i; \alpha, \epsilon, S) + \alpha_w x_i + \alpha_r z(x_i, u)$$

$$: \alpha \in \mathbb{R}^{d_\theta \times |S|}, \alpha_w^{(1)} \in \mathbb{R}^{d_y \times d_x}, \alpha_r^{(2)} \in \mathbb{R}^{d_y \times d_z}, S \subseteq_{\text{fin}} \mathcal{V}[\theta, \epsilon]\},$$

where the second line follows from lemma 1. On the other hand, since the set in the first line is a subset of the set in the last line, $\{\tilde{f}_\theta(x_i; \alpha, \epsilon, S) : \alpha \in \mathbb{R}^{d_\theta \times |S|}, S \subseteq_{\text{fin}} \mathcal{V}[\theta, \epsilon]\} = \{\tilde{f}_\theta(x_i; \alpha, \epsilon, S) + \alpha_w x_i + \alpha_r z(x_i, u) : \alpha \in \mathbb{R}^{d_\theta \times |S|}, \alpha_w^{(1)} \in \mathbb{R}^{d_y \times d_x}, \alpha_r^{(2)} \in \mathbb{R}^{d_y \times d_z}, S \subseteq_{\text{fin}} \mathcal{V}[\theta, \epsilon]\}$. This immediately implies the desired statement from theorem 2. $\square$

**A.5 Proof of Theorem 4.** As shown in the proof of theorem 4, thanks to theorem 2 and lemma 1, the remaining task to prove theorem 4 is to find a set $S' \subseteq_{\text{fin}} \mathcal{V}[\theta, \epsilon]$ such that $\{\tilde{f}_\theta(x_i; \alpha', \epsilon, S') : \alpha' \in \mathbb{R}^{d_\theta \times |S'|}\} \supseteq \{\sum_{l=t}^{H} \alpha_h^{(l+1)} h^{(l)}(x_i; u) : \alpha_h \in \mathbb{R}^{d_t}\}$. Let $M^{(l')} \cdots M^{(l+1)} M^{(l)} = I$ if $l > l'$.

**Proof of Theorem 4.** Since $f$ is specified by equation 4.3 and, hence,

$$f(x; \theta) = (\partial_{\text{vec}(W^{(H+1)})} f(x; \theta)) \text{vec}(W^{(H+1)}),$$

assumption 2 is satisfied. Let $t \in \{0, \ldots, H\}$ be fixed. Let $\theta \in (\Theta_{d_y, t} \setminus \tilde{\Omega})$ be an arbitrary local minimum of $L$. Then from theorem 2, there exists $\epsilon_0 > 0$ such that for any $\epsilon \in [0, \epsilon_0)$, $L(\theta) = \inf_{S \subseteq_{\text{fin}} \mathcal{V}[\theta, \epsilon], \alpha \in \mathbb{R}^{d_\theta \times |S|}} \sum_{i=1}^{m} \lambda_i \ell(\tilde{f}_\theta(x_i; \alpha, \epsilon, S), y_i)$, where $\tilde{f}_\theta(x_i; \alpha, \epsilon, S) = \sum_{k=1}^{d_\theta} \sum_{j=1}^{|S|} \alpha_{k,j} \partial_k f_{x_i}(\theta + \epsilon S_j)$.

Let $J = \{J^{(t+1)}, \ldots, J^{(H+1)}\} \in \mathcal{J}_{n,t}[\theta]$ be fixed. Without loss of generality, for simplicity of notation, we can permute the indices of the units of each layer such that $J^{(t+1)}, \ldots, J^{(H+1)} \supseteq \{1, \ldots, d_y\}$. Let $\tilde{B}(\theta, \epsilon_1) = B(\theta, \epsilon_1) \cap \{\theta' \in \mathbb{R}^{d_\theta} : W_{i,j}^{(l+1)} = 0$ for all $l \in \{t+1, \ldots, H-1\}$ and all $(i, j) \in (\{1, \ldots, d_{l+1}\} \setminus J^{(l+1)}) \times J^{(l)}\}$. Because of the definition of the set $J$, in $\tilde{B}(\theta, \epsilon_1)$ with $\epsilon_1 > 0$ being sufficiently small, we have that for any $l \in \{t, \ldots, H\}$,

$$f_{x_i}(\theta) = A^{(H+1)} \cdots A^{(l+2)}[A^{(l+1)} \quad C^{(l+1)}]h^{(l)}(x_i; \theta) + \varphi_{x_i}^{(l)}(\theta),$$

where

$$\varphi_{x_i}^{(l)}(\theta) = \sum_{l'=l}^{H-1} A^{(H+1)} \cdots A^{(l'+3)} C^{(l'+2)} \tilde{h}^{(l'+1)}(x_i; \theta)$$

and

$$\tilde{h}^{(l)}(x_i; \theta) = \sigma^{(l)}(B^{(l)} \tilde{h}^{(l-1)}(x_i; \theta)),$$

for all $l \geq t+2$ with $\tilde{h}^{(t+1)}(x_i; \theta) = \sigma^{(t+1)}([\xi^{(l)} \quad B^{(l)}] h^{(t)}(x_i; \theta))$. Here,

$$\begin{bmatrix} A^{(l)} & C^{(l)} \\ \xi^{(l)} & B^{(l)} \end{bmatrix} = W^{(l)}$$

with $A^{(l)} \in \mathbb{R}^{d_y \times d_y}$, $C^{(l)} \in \mathbb{R}^{d_y \times (d_{l-1} - d_y)}$, $B^{(l)} \in \mathbb{R}^{(d_l - d_y) \times (d_{l-1} - d_y)}$, and $\xi^{(l)} \in \mathbb{R}^{(d_l - d_y) \times d_y}$. Let $\epsilon_1 > 0$ be a such number, and let $\epsilon \in (0, \min(\epsilon_0, \epsilon_1/2))$ be fixed so that both the equality from theorem 2 and the above form of $f_{x_i}$ hold in $\tilde{B}(\theta, \epsilon)$. Let $R^{(l)} = [A^{(l)} \quad C^{(l)}]$.

We will now find sets $S^{(t)}, \ldots, S^{(H)} \subseteq_{\text{fin}} \mathcal{V}[\theta, \epsilon]$ such that

$$\{\tilde{f}_\theta(x_i; \alpha, \epsilon, S^{(l)}) : \alpha \in \mathbb{R}^{d_\theta}\} \supseteq \{\alpha_h^{(l+1)} h^{(l)}(x_i; u) : \alpha_h^{(l+1)} \in \mathbb{R}^{d_y \times d_l}\}.$$

Find $S^{(l)}$ with $l = H$: Since

$$(\partial_{\text{vec}(R^{(H+1)})} f_{x_i}(\theta)) \text{vec}(\alpha_h^{(H+1)}) = \alpha_h^{(H+1)} h^{(H)}(x_i; \theta),$$

$S^{(H)} = \{0\} \subseteq_{\text{fin}} \mathcal{V}[\theta, \epsilon]$ (where $0 \in \mathbb{R}^{d_\theta}$) is the desired set.

Find $S^{(l)}$ with $l \in \{t, \ldots, H-1\}$: With $\alpha_r^{(l+1)} \in \mathbb{R}^{d_{l+1} \times d_l}$, we have that

$$(\partial_{\text{vec}(R^{(l+1)})} f_{x_i}(\theta)) \text{vec}(\alpha_r^{(l+1)}) = A^{(H+1)} \cdots A^{(l+2)} \alpha_r^{(l+1)} h^{(l)}(x_i; \theta).$$

Therefore, if $\text{rank}(A^{(H+1)} \cdots A^{(l+2)}) \geq d_y$, since $\{A^{(H+1)} \cdots A^{(l+2)} \alpha_r^{(l+1)} : \alpha_r^{(l+1)} \in \mathbb{R}^{d_{l+1} \times d_l}\} \supseteq \{\alpha_h^{(l+1)} \in \mathbb{R}^{d_y \times d_l}\}$, $S^{(l)} = \{0\} \subseteq_{\text{fin}} \mathcal{V}[\theta, \epsilon]$ (where $0 \in \mathbb{R}^{d_\theta}$) is the desired set. Let us consider the remaining case: let rank($A^{(H+1)} \cdots A^{(l+2)}$) < $d_y$ and let $l \in \{t, \ldots, H-1\}$ be fixed. Let $l^* = \min\{l' \in \mathbb{Z}^+ : l+3 \le l' \le H+2 \land \text{rank}(A^{(H+1)} \cdots A^{(l')}) \ge d_y\}$, where $A^{(H+1)} \cdots A^{(H+2)} \triangleq I_{d_y}$ and the minimum exists since the set is finite and contains at least $H+2$ (nonempty). Then rank($A^{(H+1)} \cdots A^{(l^*)}$) $\ge d_{H+1}$ and rank($A^{(H+1)} \cdots A^{(l')}$) < $d_{H+1}$ for all $l' \in \{l+2, l+3, \ldots, l^*-1\}$. Thus, for all $l' \in \{l+1, l+2, \ldots, l^*-2\}$, there exists a vector $a_{l'} \in \mathbb{R}^{d_y}$ such that

$$a_{l'} \in \text{Null}(A^{(H+1)} \cdots A^{(l'+1)}) \text{ and } \|a_{l'}\|_2 = 1.$$

Let $a_{l'}$ denote such a vector. Consider $S^{(l)}$ such that the weight matrices $W$ are perturbed with $\bar{\theta} + \epsilon S_j^{(l)}$ as

$$\tilde{A}_j^{(l')} = A^{(l')} + \epsilon a_{l'} b_{l',j}^\top \text{ and } \tilde{R}_j^{(l+1)} = R^{(l+1)} + \epsilon a_{l+1} b_{l+1,j}^\top$$

for all $l' \in \{l+2, l+3, \ldots, l^*-2\}$, where $\|b_{l',j}\|_2$ is bounded such that $\|S_j^{(l)}\|_2 \le 1$. That is, the entries of $S_j$ are all zeros except the entries corresponding to $A^{(l')}$ (for $l' \in \{l+2, l+3, \ldots, l^*-2\}$) and $R^{(l+1)}$. Then $S_j^{(l)} \in \mathcal{V}[\theta, \epsilon]$, since $A^{(H+1)} \cdots A^{(l'+1)} \tilde{A}_j^{(l')} = A^{(H+1)} \cdots A^{(l'+1)} A^{(l')}$ for all $l' \in \{l+2, l+3, \ldots, l^*-2\}$ and $A^{(H+1)} \cdots A^{(l+2)} \tilde{R}_j^{(l+1)} = A^{(H+1)} \cdots A^{(l+2)} R^{(l+1)}$. Let $|S^{(l)}| = 2N$ with some integer $N$ to be chosen later. Define $S_{j+N}^{(l)}$ for $j = 1, \ldots, N$ by setting $S_{j+N}^{(l)} = S_j^{(l)}$ except that $b_{l+1,j+N} = 0$ whereas $b_{l+1,j}$ is not necessarily zero. By setting $\alpha_{j+N} = -\alpha_j$ for all $j \in \{1, \ldots, N\}$, with $\alpha_j \in \mathbb{R}^{d_{l^*} \times d_{l^*-1}}$,

$$\tilde{f}_\theta(x_i; \alpha, \epsilon, S^{(l)})$$
$$= \sum_{j=1}^N A^{(H+1)} \cdots A^{(l^*)}(\alpha_j + \alpha_{j+N}) \tilde{A}^{(l^*-2)} \cdots \tilde{A}^{(l+2)} R^{(l+1)} h^{(l)}(x_i; \theta)$$
$$+ \sum_{j=1}^N (\partial_{\text{vec}(A^{(l^*-1)})} \varphi_{x_i}^{(l)}(\theta + \epsilon S_j)) \text{vec}(\alpha_j + \alpha_{j+N})$$
$$+ \epsilon \sum_{j=1}^N A^{(H+1)} \cdots A^{(l^*)} \alpha_j \tilde{A}^{(l^*-2)} \cdots \tilde{A}^{(l+2)} a_{l+1} b_{l+1,j}^\top h^{(l)}(x_i; \theta)$$
$$= \epsilon \sum_{j=1}^N A^{(H+1)} \cdots A^{(l^*)} \alpha_j \tilde{A}^{(l^*-2)} \cdots \tilde{A}^{(l+2)} a_{l+1} b_{l+1,j}^\top h^{(l)}(x_i; \theta),$$

where we used the fact that $\partial_{\text{vec}(A^{(l^*-1)})}\varphi_{x_i}^{(l)}(\theta + \epsilon S_j)$ does not contain $b_{l+1,j}$. Since $\text{rank}(A^{(H+1)} \cdots A^{(l^*)}) \geq d_y$ and $\{A^{(H+1)} \cdots A^{(l^*)}\alpha_j : \alpha_j \in \mathbb{R}^{d_{l^*} \times d_{l^*-1}}\} = \{\frac{1}{\epsilon}\alpha'_j : \alpha'_j \in \mathbb{R}^{d_y \times d_{l^*-1}}\}$, we have that $\forall \alpha'_j \in \mathbb{R}^{d_y \times d_{l^*-1}}, \exists \alpha \in \mathbb{R}^{d_\theta \times |S|}$,

$$\tilde{f}_\theta(x_i; \alpha, \epsilon, S^{(l)}) = \sum_{j=1}^{N} \alpha'_j \tilde{A}^{(l^*-2)} \cdots \tilde{A}^{(l+2)} a_{l+1} b_{l+1,j}^\top h^{(l)}(x_i; \theta).$$

Let $N = 2N_1$. Define $S_{j+N_1}^{(l)}$ for $j = 1, \ldots, N_1$ by setting $S_{j+N_1}^{(l)} = S_j^{(l)}$ except that $b_{l^*-2, j+N_1} = 0$, whereas $b_{l^*-2, j}$ is not necessarily zero. By setting $\alpha'_{j+N_1} = -\alpha'_j$ for all $j \in \{1, \ldots, N_1\}$,

$$\tilde{f}_\theta(x_i; \alpha, \epsilon, S^{(l)}) = \epsilon \sum_{j=1}^{N_1} \alpha'_j a_{l^*-2} b_{l^*-2, j}^\top \tilde{A}^{(l^*-3)} \cdots \tilde{A}^{(l+2)} a_{l+1} b_{l+1,j}^\top h^{(l)}(x_i; \theta).$$

By induction,

$$\tilde{f}_\theta(x_i; \alpha, \epsilon, S^{(l)}) = \epsilon^t \sum_{j=1}^{N_t} \alpha'_j a_{l^*-2} b_{l^*-2, j} a_{l^*-3} b_{l^*-3, j} \cdots a_{l+1} b_{l+1, j}^\top h^{(l)}(x_i; \theta),$$

where $t = (l^* - 2) - (l + 2) + 1$ is finite. By setting $\alpha'_j = \frac{1}{\epsilon^t} q_j a_{l^*-2}^\top$ and $b_{l,j} = a_{l-1}$ for all $l = l^* - 2, \ldots, l$ ($\epsilon > 0$),

$$\tilde{f}_\theta(x_i; \alpha, \epsilon, S^{(l)}) = \sum_{j=1}^{N_t} q_j b_{l+1,j}^\top h^{(l)}(x_i; \theta).$$

Since $q_j b_{l+1,j}$ are arbitrary, with sufficiently large $N_t$ ($N_t = d_y d_l$ suffices), we can set $\sum_{j=1}^{N_t} q_j b_{l+1,j} = \alpha_h^{(l)}$ for any $\alpha_h^{(l)} \in \mathbb{R}^{d_\theta \times d_l}$, and hence

$$\{\tilde{f}_\theta(x_i; \alpha, \epsilon, S^{(l)}) : \alpha \in \mathbb{R}^{d_\theta \times |S^{(l)}|}\} \supseteq \{\alpha_h^{(l)} h^{(l)}(x_i; \theta) : \alpha_h^{(l)} \in \mathbb{R}^{d_\theta \times d_l}\}.$$

Thus far, we have found the sets $S^{(t)}, \ldots, S^{(H)} \subseteq_{\text{fin}} \mathcal{V}[\theta, \epsilon]$ such that $\{\tilde{f}_\theta(x_i; \alpha, \epsilon, S^{(l)}) : \alpha \in \mathbb{R}^{d_\theta}\} \supseteq \{\alpha_h^{(l+1)} h^{(l)}(x_i; u) : \alpha_h^{(l+1)} \in \mathbb{R}^{d_y \times d_l}\}$. From lemma 1, we can combine these, yielding

$$\{\tilde{f}_\theta(x_i; \alpha, \epsilon, S) : \alpha \in \mathbb{R}^{d_\theta}, S \subseteq_{\text{fin}} \mathcal{V}[\theta, \epsilon]\}$$
$$= \left\{ \sum_{l=t}^{H} \tilde{f}_\theta(x_i; \alpha^{(l)}, \epsilon, S^{(l)}) + \tilde{f}_\theta(x_i; \alpha, \epsilon, S) : \alpha^{(t)}, \ldots, \alpha^{(H)} \in \mathbb{R}^{d_\theta}, \right.$$

$$\alpha \in \mathbb{R}^{d_\theta}, S \subseteq_{\text{fin}} \mathcal{V}[\theta, \epsilon]\}$$

$$\supseteq \left\{ \sum_{l=t}^{H} \alpha_h^{(l+1)} h^{(l)}(x_i; u) + \tilde{f}_\theta(x_i; \alpha, \epsilon, S) : \alpha_h^{(l+1)} \in \mathbb{R}^{d_y \times d_l}, \right.$$

$$\left. \alpha \in \mathbb{R}^{d_\theta \times |S|}, S \subseteq_{\text{fin}} \mathcal{V}[\theta, \epsilon] \right\}.$$

Since the set in the first line is a subset of the set in the last line, the equality holds in the above equation. This immediately implies the desired statement from theorem 2. □

## Acknowledgments

We gratefully acknowledge support from NSF grants 1523767 and 1723381, AFOSR grant FA9550-17-1-0165, ONR grant N00014-18-1-2847, Honda Research, and the MIT-Sensetime Alliance on AI. Any opinions, findings, and conclusions or recommendations expressed in this material are our own and do not necessarily reflect the views of our sponsors.